\pdfoutput=1
\documentclass{article}
\usepackage{geometry}
\newgeometry{
    textheight=9in,
    textwidth=5.5in,
    top=1in,
    headheight=12pt,
    headsep=25pt,
    footskip=30pt
  }

\usepackage[T1]{fontenc}
\usepackage{microtype}
\usepackage[hyperref]{xcolor}
\usepackage[hang, flushmargin]{footmisc}

\usepackage{math}
\usepackage{xkcd_color}

\usepackage[
    colorlinks=true,
    linkcolor={xkcd red},
    citecolor={xkcd blue},
    urlcolor={xkcd magenta},
    backref=section,
]{hyperref}
\usepackage{footnotebackref}
\usepackage[capitalise]{cleveref}

\usepackage{natbib}
\usepackage{graphicx}
\graphicspath{{./figures/}}
\usepackage{subcaption}

\usepackage{booktabs}
\usepackage{multirow}
\usepackage{listings}
\usepackage[inline]{enumitem}

\usepackage{siunitx}
\usepackage{amssymb}

\newcommand{\KL}{{\mathrm{KL}}}
\newcommand{\Yt}{{\widetilde{Y}}}
\newcommand{\yt}{{\widetilde{y}}}

\title{%
Approximating Instance-Dependent Noise\\
via Instance-Confidence Embedding
}

\author{%
Yivan Zhang$^{1,2}$ \and 
Masashi Sugiyama$^{2,1}$
}
\date{%
$^1$The University of Tokyo\\
$^2$RIKEN AIP
}

\begin{document}

\maketitle

\begin{abstract}
Label noise in multiclass classification is a major obstacle to the deployment of learning systems.
However, unlike the widely used \emph{class-conditional noise} (CCN) assumption that the noisy label is independent of the input feature given the true label, label noise in real-world datasets can be aleatory and heavily dependent on individual instances.
In this work, we investigate the \emph{instance-dependent noise} (IDN) model and propose an efficient approximation of IDN to capture the instance-specific label corruption.
Concretely, noting the fact that most columns of the IDN transition matrix have only limited influence on the class-posterior estimation, we propose a variational approximation that uses a single-scalar \emph{confidence} parameter.
To cope with the situation where the mapping from the instance to its confidence value could vary significantly for two adjacent instances, we suggest using \emph{instance embedding} that assigns a trainable parameter to each instance.
The resulting \emph{instance-confidence embedding} (ICE) method not only performs well under label noise but also can effectively detect ambiguous or mislabeled instances.
We validate its utility on various image and text classification tasks.
\end{abstract}

\section{Introduction}
\label{sec:introduction}
In modern machine learning, large-scale data has become indispensable \citep{russakovsky2015imagenet, wang2018glue}.
A prevalent approach to collecting large-scale labeled datasets is to use imperfect sources such as crowdsourcing and web crawling \citep{fergus2005learning, schroff2010harvesting, wang2018glue}, which is usually less expensive and time-consuming than manual annotation by domain experts.
However, such methods inevitably introduce label noise that may lead to overfitting and hurt the generalization of deep models \citep{arpit2017closer, zhang2017understanding}.


In such situations, it is often beneficial to 
(i) remove mislabeled data or abstain from using confusing instances \citep{hara2019data, thulasidasan2019combating};
(ii) increase robustness and reduce harmful influences of noisy labels \citep{malach2017decoupling, mirzasoleiman2020coresets, liu2020early};
or (iii) explicitly model the transition from the unobservable true label to the noisy observation \citep{goldberger2017training, patrini2017making, xia2020part}.
In this work, we focus on explicit modeling of the label corruption process, which is model-agnostic and data-efficient.


Most existing studies in this direction employ the \emph{class-conditional noise} (CCN) assumption, i.e., the noisy label is independent of the input feature given the true label \citep{angluin1988learning, natarajan2013learning, patrini2017making}.
However, this assumption could be too strong to fit some real-world data well \citep{xiao2015learning, chen2021beyond}.
More importantly, CCN only captures the general label flipping patterns between classes for all instances.
In applications such as data cleansing and human-in-the-loop interaction, instance-specific noise information itself could be of central interest.
This urges us to consider not only the class-conditional noise pattern but also the instance-specific noise modeling.


To handle this problem, in this work, we study the \emph{instance-dependent noise} (IDN) model, where the noisy label also depends on the input.
Several methods have been reported in the literature, but they either only focus on binary classification under strong assumptions \citep{menon2018learning, cheng2020learning} or are based on domain-specific knowledge \citep{xia2020part}.
In contrast, we propose a simple domain-agnostic approximation method for the multiclass IDN model, referred to as \emph{instance-confidence embedding} (ICE).
Concretely, to avoid estimating a noise transition matrix for each instance, we propose a variational approximation that uses a scalar \emph{confidence} parameter (\cref{ssec:approximation}).
Then, we suggest to use \emph{instance embedding} that assigns a trainable parameter to each instance because the mapping from the instance to its confidence value could be non-smooth and is usually not required to generalize to unseen examples (\cref{ssec:embedding}).
Lastly, we show the effectiveness of the proposed method and its ability to detect ambiguous or mislabeled instances through experiments on various image and text classification tasks (\cref{sec:ex}).

\section{Problem: Instance-Dependent Noise}
\label{sec:problem}
\begin{figure}[t]
\centering
\begin{subfigure}{.24\linewidth}
\centering
\resizebox{.8\linewidth}{!}{%
\begin{tikzpicture}
\platenotation
\node (X) [observable] {$X$};
\node (Y) [unobservable, below = of X] {$Y$};
\node (Yt) [observable, right = of Y] {$\Yt$};
\path (X) edge [dependency] (Y);
\path (X) edge [dependency] (Yt);
\end{tikzpicture}
}
\caption{IND}
\end{subfigure}
\begin{subfigure}{.24\linewidth}
\centering
\resizebox{.8\linewidth}{!}{%
\begin{tikzpicture}
\platenotation
\node (X) [observable] {$X$};
\node (Y) [unobservable, below = of X] {$Y$};
\node (Yt) [observable, right = of Y] {$\Yt$};
\path (X) edge [dependency] (Y);
\path (Y) edge [dependency] (Yt);
\end{tikzpicture}
}
\caption{CCN}
\end{subfigure}
\begin{subfigure}{.24\linewidth}
\centering
\resizebox{.8\linewidth}{!}{%
\begin{tikzpicture}
\platenotation
\node (X) [observable] {$X$};
\node (Y) [unobservable, below = of X] {$Y$};
\node (Yt) [observable, right = of Y] {$\Yt$};
\path (X) edge [dependency] (Y);
\path (Y) edge [dependency] (Yt);
\path (X) edge [dependency] (Yt);
\end{tikzpicture}
}
\caption{IDN}
\end{subfigure}
\begin{subfigure}{.24\linewidth}
\centering
\resizebox{.8\linewidth}{!}{%
\begin{tikzpicture}
\platenotation
\node (X) [observable] {$X$};
\node (Y) [unobservable, below = of X] {$Y$};
\node (Yt) [observable, right = of Y] {$\Yt$};
\node (C) [unobservable, above = of Yt] {$C$};
\path (X) edge [dependency] (Y);
\path (Y) edge [dependency] (Yt);
\path (X) edge [dependency] (C);
\path (C) edge [dependency] (Yt);
\end{tikzpicture}
}
\caption{ICE}
\end{subfigure}
\caption{%
\textbf{Graphical representations} of noise models, including the the conditionally independent labels (IND) model, class-conditional noise (CCN) model, instance-dependent noise (IDN) model, and the proposed \emph{instance-confidence embedding} (ICE) approximation of IDN.
Here, $X$ is the \emph{input feature}, $Y$ is the \emph{true label}, $\Yt$ is the \emph{noisy label}, and $C \in [0, 1]$ is a scalar \emph{confidence} parameter.
}
\label{fig:plate}
\end{figure}
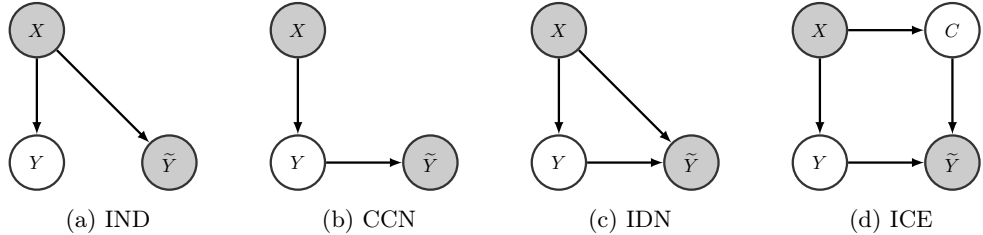


In this section, we give a brief overview of learning with \emph{instance-dependent noise} (IDN).


\subsection{Notation}

Consider a $K$-class classification problem, where $X \in \sX$ is the \emph{input feature} and $Y \in \{1, \dots, K\}$ is the unobservable \emph{true label}.
We assume that the \emph{clean class-posterior} $p(Y | X)$ comes from a parametric family of distributions:
\begin{equation}
\label{eq:classifier}
  p_\phi(Y | X) 
\defeq \Categorical(Y | \vp = f(X; \phi))
,
\end{equation}
where $\vp \in \Delta^{K-1}$ is the probability parameter for $Y$ in the $(K-1)$-dimensional probability simplex $\Delta^{K-1}$, and $f: \sX \to \Delta^{K-1}$ is a differentiable function parameterized by $\phi$ that maps the feature $X$ to its corresponding probability parameter $\vp$.
Then, let $\Yt \in \{1, \dots, K\}$ be the \emph{noisy label}.
The goal is to predict $Y$ from $X$ based on a finite i.i.d.~sample of $(X, \Yt)$-pairs.


\subsection{Dependence}

Next, we introduce the dependence structure between $X$, $Y$, and $\Yt$, which characterize different noise models.
The graphical representations of noise models are illustrated in \cref{fig:plate}.

In IDN, we assume that the joint distribution of $X$, $Y$, and $\Yt$ can be factorized as follows:
\begin{equation}
  p(X, Y, \Yt)
= p(\Yt | Y, X) p_\phi(Y | X) p(X)
.
\end{equation}
That is, the noisy label $\Yt$ depends on both the instance $X$ and the true label $Y$.
Then, the \emph{noisy class-posterior} $p(\Yt | X)$ can be obtained by marginalizing $p(Y, \Yt | X)$ over $Y$:
\begin{equation}
\label{eq:idn}
\textstyle
  p_\phi(\Yt | X)
\defeq \Categorical(\Yt | \vq = \sum_{Y=1}^K p(\Yt | Y, X) p_\phi(Y | X))
,
\end{equation}
where $\vq \in \Delta^{K-1}$ denotes the probability parameter for $\Yt$.

Note that $p(\Yt | Y, X)$ plays a central role in IDN.
Since both $Y$ and $\Yt$ are categorical random variables, for a certain instance $x$, $p(\Yt | Y, X=x)$ can be seen as a $K \times K$ stochastic matrix $\mT(x)$, whose elements are $\mT_{ij}(x) \defeq p(\Yt=j | Y=i, X=x)$ for $i, j \in \{1, \dots, K\}$.
Conventionally, $\mT(x)$ is called a \emph{noise transition matrix} \citep{patrini2017making}.
Then, $p(\Yt | Y, X)$ can be regarded as a matrix-valued function $\mT: \sX \to [0, 1]^{K \times K}$ that maps each instance $x$ to its corresponding IDN transition matrix $\mT(x)$.
Without any restriction, we need $K \times K$ parameters for each instance $x$.


\subsection{Approach}
\label{ssec:approach}

Owing to its complexity, IDN has only been studied to a limited extent but is of great interest recently.
A straightforward method is to jointly estimate the matrix-valued function $\mT(x)$ as well as the clean class-posterior $p_\phi(Y | X)$ using neural networks \citep{goldberger2017training}.
However, the estimation error of $\mT(x)$ could be high, which deteriorates the classification performance.
Another direction is to restrict the problem under certain conditions, so that we can provide theoretical guarantees \citep{menon2018learning, cheng2020learning}.
However, existing work mainly focused on binary classification.

A promising approach is to approximate IDN using a simpler dependence structure, such as a mixture of noises with different semantic meanings \citep{xiao2015learning} or a weighted combination of noises that depend on parts of the instance \citep{xia2020part}.
In this work, we also suggest that it might be unnecessary to obtain a $K \times K$ matrix for each instance $x$:
Note that $p_\phi(\Yt | x)$ can be seen as a linear combination of columns of $\mT(x)$ weighted by $p_\phi(Y | x)$;
If the maximum value of $p_\phi(Y | x)$ is close to $1$, i.e., the label of the instance $x$ is almost deterministic, the estimation of $K - 1$ columns of $\mT(x)$ has only limited influence on the estimated noisy class-posterior $\widehat{p}(\Yt | x)$.
This suggests the possibility of using a relatively simple model to approximate $p(\Yt | X)$ in real-world applications.
In this work, we consider a \emph{single-parameter} approximation for each instance, which is introduced in \cref{ssec:approximation} and illustrated in \cref{fig:noise_approx}.

Another issue is that existing methods still introduce some level of smoothness w.r.t.~$x$ into $\mT(x)$ \citep{goldberger2017training, xiao2015learning, xia2020part}.
In real-world problems, however, we can only access a finite sample of $(X, \Yt)$-pairs that are possibly annotated by non-experts or web crawlers \citep{fergus2005learning}.
Thus, the label noise could be aleatory and $\mT(x)$ could vary significantly for two adjacent instances.
Also, the classifier $p_\phi(Y | X)$ is desired but the generalization of $\mT(x)$ to unseen examples is usually dispensable.
This inspires us to use \emph{instance embedding} instead of neural network approximation, which is discussed in \cref{ssec:embedding} and demonstrated in \cref{fig:demo}.

\section{Proposed Method}
\label{sec:method}
In this section, we present our proposed method, \emph{instance-confidence embedding} (ICE).


\subsection{Variational lower bound}

Note that $\mT(x)$ serves as a \emph{linear mapping} from $\vp$ to $\vq$ (\cref{eq:idn}).
Due to the difficulty of estimating the matrix-valued function $\mT(x)$, we use a simpler function $q_{\theta, \phi}(\Yt | X)$ parameterized by $\theta$ as a \emph{variational approximation} to $p_\phi(\Yt | X)$.
The choice of the approximation family is discussed in Section~\ref{ssec:approximation}.

Then, let us consider the \emph{expected log-likelihood} as the learning objective, which can be rewritten as
\begin{equation}
\label{eq:likelihood}
  \E_{\Yt \sim p(\Yt | X)} 
  \brackets{\log p(\Yt | X)}
= 
  D_\KL \diver{p_\phi(\Yt | X)}{q_{\theta, \phi}(\Yt | X)}
+ \mathcal{L}(\theta, \phi; X)
,
\end{equation}
where $D_\KL$ denotes the Kullback-Leibler (KL) divergence, and the second term is
\begin{equation}
  \mathcal{L}(\theta, \phi; X)
\defeq
  \E_{\Yt \sim p(\Yt | X)} 
  \brackets{\log q_{\theta, \phi}(\Yt | X)}
.
\end{equation}
Since the KL-divergence is always non-negative, this term gives a \emph{variational lower bound} of the expected log-likelihood.
Then, we have the following learning objective to maximize:
\begin{equation}
\label{eq:objective}
  L(\theta, \phi)
\defeq
  \E_{X \sim p(X)} 
  \brackets{\mathcal{L}(\theta, \phi; X)}
= 
  \E_{X, \Yt \sim p(X, \Yt)} 
  \brackets{\log q_{\theta, \phi}(\Yt | X)}
.
\end{equation}
In practice, the expectation can be approximated using the empirical distribution based on a finite i.i.d.~sample of $(X, \Yt)$-pairs.


\begin{figure}[t]
\centering
\includegraphics[width=\linewidth]
{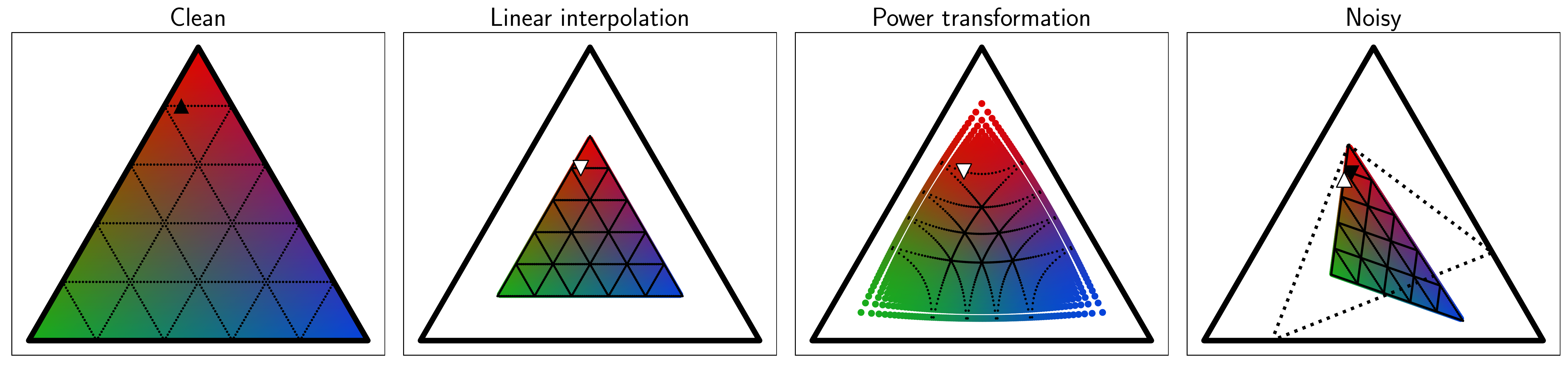}
\caption{%
An illustration of the transformation ($\blacktriangle \mapsto \blacktriangledown$) from the clean class-posterior $p_\phi(Y | x)$ (the leftmost) to the noisy class-posterior $p_\phi(\Yt | x)$ (the rightmost).
We can see that when the label is almost deterministic ($\blacktriangle$ is close to a vertex), the estimation of $K - 1$ columns of the transition matrix $\mT(x)$ (the two deviated vertices of the dotted triangle) has only limited influence on the estimated noisy class-posterior $\widehat{p}(\Yt | x)$ ($\vartriangle$ is still close to $\blacktriangledown$).
This inspires us to go a step further and use \textbf{single-parameter approximations} ($\blacktriangle \mapsto \triangledown$) $q_{\theta,\phi}(\Yt | x)$ (\cref{eq:linear,eq:power}).
}
\label{fig:noise_approx}
\end{figure}


\subsection{Variational approximation}
\label{ssec:approximation}

Next, we discuss the choice of the variational approximation family of
$q_{\theta, \phi}(\Yt | X)$.

To approximate the effect of multiplying an IDN transition matrix $\mT(x)$ that requires $K \times K$ parameters for each instance $x$, in this work, we use a simpler transformation from $\vp$ to $\vq$, which is not necessarily linear.
Compared with estimating a full matrix for each instance without any restriction \citep{goldberger2017training}, obtaining only an approximation may cause higher approximation error, but on the other hand, it may reduce estimation error and thus improve the classification performance. 
The high estimation error of complex models might be more harmful, which is empirically validated in \cref{sec:ex}.
It is also the case when using CCN as an approximation of IDN to balance this trade-off.
The difference is that CCN obtains a \emph{complete} transition matrix for \emph{all} instances, but ICE aims to obtain an \emph{approximated} trend for \emph{each} instance, which gives useful instance-specific noise information.

Then, we suggest to use a single-scalar parameter $C \in [0, 1]$ for each instance to control this approximation, which is useful for sorting and comparing training examples.
This parameter is referred to as the \emph{confidence} and is obtained via a function $g: \sX \to [0, 1]$ parameterized by $\theta$, i.e., $C = g(X; \theta)$.
The confidence $C$ plays a central role in our method, where $C = 0$ means that the instance is ambiguous or mislabeled and thus the classifier should not give a confident prediction.

Finally, we need to design a transformation from $\vp$ to $\vq$ parameterized by the confidence $C$.
We denote this function by $h: \Delta^{K-1} \to \Delta^{K-1}$.
In summary, $q_{\phi, \theta}(\Yt | X)$ takes the following form:
\begin{equation}
  q_{\phi, \theta}(\Yt | X)
\defeq 
  \Categorical(\vq = h(f(X; \phi); g(X; \theta)))
.
\end{equation}

Next, we analyze what characteristics $h$ needs to have.
First, we suggest that $h$ needs not necessarily to be a linear transformation because the transformation is instance-dependent and any function that maps $\vp$ to $\vq$ as close as possible for a certain instance $x$ would suffice.
Second, we require that $\argmax(\vp) = \argmax(\vq)$, i.e., $\vq = h(\vp; C)$ should be an \emph{argmax-preserving} function so that the top-1 index of the probability vector does not change.
This is because $h$ should only affect the confidence of the prediction, not the final decision.
Otherwise, if $h$ is too flexible and is able to map a confident prediction to a different confident prediction, then the output of $f$ could be arbitrary, and consequently, no information of the true label can be learned from the noisy label supervision.

Based on this motivation and the aforementioned semantics of $C$, we require that $h(\vp; 1) = \vp$ and $h(\vp; 0) = \vu$, where $\vu \in \Delta^{K-1}$ is the uniform probability vector ($\vu_i = \frac1K$).
Then, when the confidence $C$ is high, the classifier gives a prediction closer to the original \emph{confident prediction} $\vp$; and when the confidence $C$ is low, the classifier tends to give a \emph{random guess} $\vu$.

Here, we propose two functions for $h$ that satisfy the above conditions:
\begin{align}
\label{eq:linear}
\phantom{\text{(power transformation)}}
  \vq_i &= C \vp_i + (1 - C) \vu_i
,
& \text{(linear interpolation)}
\\
\label{eq:power}
  \vq_i &= \frac{\vp_i^C}{\sum_{j=1}^K \vp_j^C}
,
& \text{(power transformation)}
\end{align}
for $i = 1, \dots, K$.
The visualization of these two transformations for $K = 3$ is given in \cref{fig:noise_approx}.


\begin{figure}[t]
\centering
\includegraphics[width=\linewidth]
{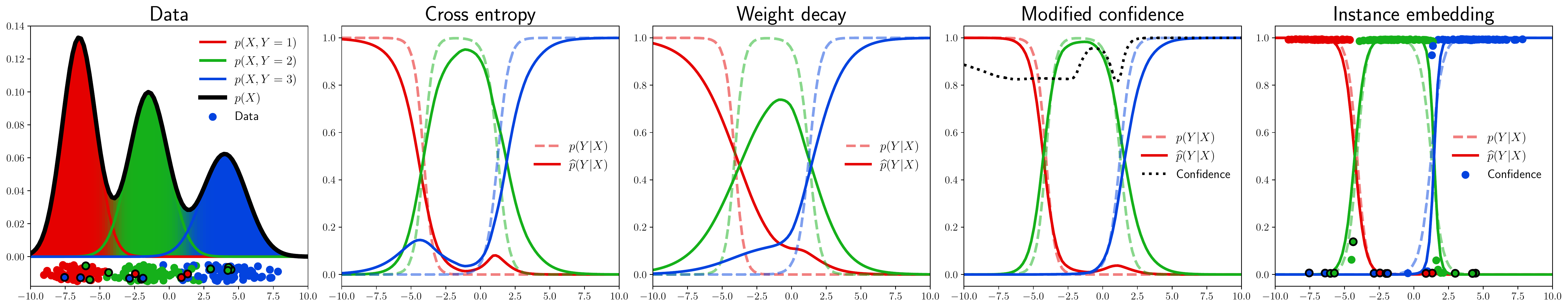}
\caption{%
An example of the learned \textbf{class-posteriors} using 
(a) the usual cross entropy without any modification;
(b) the one with weight decay as a regularization;
(c) modified confidence of the prediction (\cref{eq:linear}) with neural network approximation for $g: \sX \to [0, 1]$; and
(d) the one with instance embedding, i.e., the proposed \emph{instance-confidence embedding} (ICE) method.
The points with black edges are mislabeled instances.
We can observe that noisy labels affect the decision boundary and the model complexity.
With weight decay, the model complexity can be controlled but the confidence is deviated.
Comparing the last two panels, modifying the confidence of the prediction via \cref{eq:linear} works better with instance embedding than neural network approximation.
ICE can reduce the influence of ambiguous or mislabeled instances to improve the class-posterior estimation.
}
\label{fig:demo}
\end{figure}


\subsection{Instance embedding}
\label{ssec:embedding}

The last piece of our method is the choice of $g: \sX \to [0, 1]$,  the function that maps the instance $x$ to its confidence value $C$.
It is also possible to use a neural network to approximate this function.
However, because we usually only have a limited number of training examples and $g$ could be non-smooth w.r.t.~its input $x$, $g$ may not be well approximated by a neural network with similar complexity to the classifier $f$, which is illustrated in an example in \cref{fig:demo}.
Further, $g$ may be rarely needed after training so its generalization ability is not required in many cases.

Based on these facts, we propose to use \emph{instance embedding}, i.e., to assign a trainable parameter to each instance $x$.
In other words, the only feature for an instance we use is its \emph{index} in the training dataset.
In this way, $g$ is expressive and flexible but cannot be used for predicting the confidence of unseen instances.
Accordingly, for a training dataset of size $N$, we need $N$ parameters for a one-dimensional instance embedding.

This seems to be a high additional computational cost when the dataset size is large, but it is often acceptable, because 
(i) in modern deep learning, it is common to use over-parameterized models \citep{nakkiran2019deep}, and the number of instances is usually not comparable to the number of parameters of the classifier $f$ (e.g., CIFAR-10 \citep{krizhevsky2009learning}: $\num{5e4}$, ResNet-18 \citep{he2016deep}: $\sim \num{1e7}$); and 
(ii) the gradient of the instance embedding is sparse and only a small subset of parameters needs to be updated at each iteration.

The idea of associating an entity with a scalar or vector embedding using a simple lookup table with a fixed dictionary size has been widely used in natural language processing \citep{mikolov2013efficient, pennington2014glove, peters2018deep, devlin2018bert} due to the discrete nature of tokens, and can be seen recently in contrastive learning \citep{wu2018unsupervised, he2020momentum} for vision tasks.
Instance embedding enables the function to take any possible value on all observed instances but cannot generalize to any unseen token or image.

\section{Related Work}
\label{sec:related}
In this section, we review related problem settings and methods.


\paragraph{Class-conditional noise (CCN).}

Compared with the IDN model, the instance-independent and \emph{class-conditional noise} (CCN) model has an additional assumption: $p(\Yt | Y, X) = p(\Yt | Y)$, i.e., the noisy label $\Yt$ only depends on the true label $Y$.
CCN has been well studied in both binary \citep{angluin1988learning, long2010random, natarajan2013learning, van2015learning, liu2015classification} and multiclass \citep{patrini2017making, xia2019anchor, yao2020dual} classification.
Also, \emph{robust loss functions} \citep{ghosh2017robust, zhang2018generalized, wang2019symmetric, charoenphakdee2019symmetric, ma2020normalized, feng2020can, lyu2020curriculum, liu2020peer} have been mainly developed under the CCN setting.
In practice, CCN methods can serve as practical approximations of IDN but the assumption could be too strong to fit some real-world data well \citep{xiao2015learning}.


\paragraph{Conditionally independent labels (IND).}

The other direction is to assume $p(\Yt | Y, X) = p(\Yt | X)$, i.e., $Y$ and $\Yt$ are two sets of independent labels conditioned on the feature $X$.
This dependence structure is used in the \emph{information bottleneck} framework \citep{tishby1999information, tishby2015deep, alemi2017deep, saxe2019information}, where the learning objective is to find a representation $Y$ that is maximally informative about the observation $\Yt$ based on the mutual information.
This framework can be adapted for learning from noisy labels if we choose a categorical representation $Y$.
The graphical representations of IND and CCN are given in \cref{fig:plate}.


\paragraph{Label smoothing.}

Note that \cref{eq:linear} is similar to the \emph{label smoothing} (LS) technique \citep{szegedy2016rethinking, pereyra2017regularizing, lukasik2020does}, where the empirical distribution is linearly interpolated with a uniform distribution with a fixed mixing parameter.
It is also related to the soft/hard \emph{bootstrapping loss} \citep{reed2015training}, where the observed label is mixed with the predicted probability/predicted label.
In contrast, in our method, it is the prediction $\vp$ that is ``smoothed'', not the label.
We elucidate their relations and differences in \cref{app:lerp}.


\paragraph{Temperature scaling.}

If we use softmax as the final layer of the neural network for $p_\phi(Y | X)$, the proposed method is closely related to the \emph{temperature scaling} (TS) technique \citep{guo2017calibration}.
Concretely, if $\vp_i \propto \exp\{f_i(X; \phi)\}$ for $i = 1, \dots, K$, then \cref{eq:power} becomes
\begin{equation}
\vq_i \propto \exp\{C f_i(X; \phi)\}
,    
\end{equation}
which shows that $C$ is the \emph{reciprocal of the temperature}.
The difference is that the parameter $C$ is instance-dependent in our formulation, rather than being fixed for all instances.
Also, TS \citep{guo2017calibration} and its extensions \citep{kull2019beyond, rahimi2020intra} have been mainly used as post-hoc confidence calibration methods, while our method is used during training.


\paragraph{Sample selection.}

In a broader sense, the proposed method belongs to a category of methods that treat training examples differently in order to reduce the harmful effects of mislabeled instances.
Besides the class-posteriors that our method uses, these methods exploit the training dynamics, loss characteristics, gradient information, or information of data itself from various perspectives.
Examples include 
\emph{data cleansing} \citep{liu2008isolation, northcutt2019confident, hara2019data} that first removes harmful instances and then (re-)trains the model on the remaining subset;
\emph{dynamic training sample selection} \citep{malach2017decoupling, jiang2018mentornet, han2018co, wang2018iterative, yu2019does, mirzasoleiman2020coresets, wu2020topological, chen2021beyond} that selects training examples dynamically during training;
\emph{training techniques} \citep{menon2020can, liu2020early} that are designed to increase robustness and avoid memorization of noisy labels;
\emph{learning with rejection} or \emph{selective classification} \citep{el2010foundations, thulasidasan2019combating, mozannar2020consistent} that abstains from using confusing instances while improving the classification performance on accepted instances;
and \emph{semi-supervised learning} \citep{nguyen2019self, li2020dividemix} that exploits unlabeled data.

In the same spirit, our proposed method also attempts to detect harmful instances and reduce their influences automatically so as to improve the robustness of the class-posterior estimation.
However, unlike explicit sample selection methods, the resulting algorithm is lightweight and has a low computational cost.
Also, because the proposed method only affects the class-posterior, it is usually compatible with other training methods.
Thus, the proposed method can be used alone or integrated into an existing training pipeline to further improve the performance.

\section{Experiments}
\label{sec:ex}
In this section, we experimentally verify if the proposed \emph{instance-confidence embedding} (ICE) method is able to differentiate mislabeled instances from correct ones and consequently improve the classification performance.
We also demonstrate that there already exist ambiguous or mislabeled training examples in the original datasets which can be detected by the proposed method.
We evaluated on both image classification (\cref{ssec:image}) and text classification (\cref{ssec:text}).


\subsection{Image classification}
\label{ssec:image}

\paragraph{Datasets.}

We evaluated our method on six image classification datasets, namely
\textbf{MNIST} \citep{lecun1998gradient},
Fashion-MNIST (\textbf{FMNIST}) \citep{xiao2017fashion},
and Kuzushiji-MNIST (\textbf{KMNIST}) \citep{clanuwat2018deep} 
datasets, which contain $28\times28$ grayscale images in $10$ classes;
and \textbf{SVHN} \citep{netzer2011reading}, \textbf{CIFAR-10}, and \textbf{CIFAR-100} \citep{krizhevsky2009learning} datasets, which contain $32\times32$ colour images in $10$, $10$ and $100$ classes, respectively.

\paragraph{Methods.}

We compared the following eight methods:
\begin{enumerate*}[label=(\arabic*), itemjoin={{; }}, itemjoin*={{; and }}]
\item (\textbf{CCE})
categorical cross-entropy loss

\item (\textbf{Bootstrapping})
(hard) bootstrapping loss \citep{reed2015training} that regularizes the output with the predicted label

\item (\textbf{Adaptation})
noise adaptation layer \citep{goldberger2017training} that estimates a full $K \times K$ transition matrix for each instance

\item (\textbf{Forward})
forward correction \citep{patrini2017making} that estimates a transition matrix for all instances 

\item (\textbf{DAC})
deep abstaining classifier \citep{thulasidasan2019combating} that uses abstention for robust learning

\item (\textbf{GCE})
generalized cross-entropy loss \citep{zhang2018generalized} as a robust loss

\item (\textbf{ICE-LIN})
instance-confidence embedding with the linear interpolation (\cref{eq:linear})

\item (\textbf{ICE-POW})
the one with the power transformation (\cref{eq:power})
\end{enumerate*}.
For a fair comparison, we implemented aforementioned methods using the same network architecture and hyperparameters.

\paragraph{Models.}

For MNIST, FMNIST, and KMNIST, we used a sequential convolutional neural network (CNN) and an Adam optimizer \citep{kingma2014adam}.
For SVHN, CIFAR-10 and CIFAR-100, we used a residual network model ResNet-18 \citep{he2016deep} and a stochastic gradient descent (SGD) optimizer with momentum \citep{sutskever2013importance}.
Hyperparameters are provided in \cref{app:ex}.


\begin{table}[t]
\centering
\caption{%
\textbf{Accuracy} ($\%$) on the MNIST, FMNIST, KMNIST, SVHN, CIFAR-10, and CIFAR-100 datasets where $50\%$ of labels are randomly flipped. 
``Mean (standard deviation)'' for 10 trials are reported.
Outperforming methods are highlighted in boldface using one-tailed t-tests with a significance level of $0.05$.
}
\label{tab:image}
\resizebox{\linewidth}{!}{%
\begin{tabular}{ccccccc}
\toprule
& MNIST & FMNIST & KMNIST & SVHN & CIFAR-10 & CIFAR-100
\\
\midrule

CCE           & $94.91(0.43)$ & $85.05(0.52)$ & $80.40(1.25)$ & $71.50(1.68)$ & $68.34(0.82)$ & $47.09(0.65)$ \\
Bootstrapping & $97.30(0.28)$ & $87.24(0.36)$ & $84.21(1.01)$ & $76.62(0.97)$ & $75.97(0.45)$ & $49.56(0.42)$ \\
Adaptation    & $96.27(0.41)$ & $86.20(0.87)$ & $81.38(2.07)$ & $68.58(6.45)$ & $63.95(5.94)$ & $31.70(1.28)$ \\
Forward       & $95.09(0.56)$ & $85.51(0.45)$ & $80.76(1.29)$ & $74.43(6.42)$ & $68.28(0.62)$ & $47.92(0.31)$ \\
DAC           & $96.60(0.47)$ & $86.87(0.48)$ & $82.77(0.74)$ & $\mathbf{80.97(4.83)}$ & $71.55(0.34)$ & $47.01(0.44)$ \\
GCE           & $98.31(0.13)$ & $88.76(0.26)$ & $88.39(0.60)$ & $75.03(0.98)$ & $80.38(0.67)$ & $\mathbf{55.64(0.40)}$ \\
ICE-LIN       & $\mathbf{98.64(0.15)}$ & $\mathbf{89.41(0.18)}$ & $\mathbf{89.61(0.41)}$ & $77.51(0.75)$ & $\mathbf{82.08(0.39)}$ & $\mathbf{55.30(0.47)}$ \\
ICE-POW       & $\mathbf{98.60(0.09)}$ & $\mathbf{89.29(0.20)}$ & $89.21(0.53)$ & $\mathbf{79.91(0.96)}$ & $\mathbf{82.14(0.44)}$ & $54.31(0.48)$ \\

\bottomrule
\end{tabular}
} 
\end{table}

\begin{figure}
\centering
\includegraphics[width=\textwidth]
{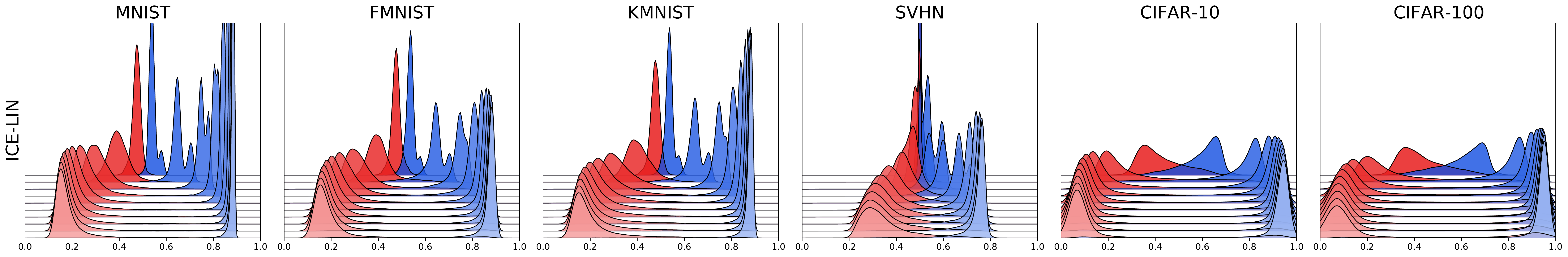}
\caption{%
\textbf{Ridgeline plots} of the confidence $C$ during training (ICE-LIN).
The density is estimated via Gaussian kernel density estimation (KDE).
The red/blue curves represent the confidence of instances with flipped/original labels, respectively.
}
\label{fig:ridge}
\end{figure}


\paragraph{Improving classification performance.}

To verify if the proposed method is able to improve the classification performance under label noise, we constructed semi-synthetic noisy datasets so that the true labels are known.
We regarded the original labels as clean labels, although as will be shown in the next experiment, label errors already exist in the original datasets to some extent.
Following a common setup \citep[e.g.,][]{reed2015training, patrini2017making, thulasidasan2019combating}, we simply flipped a fraction of labels randomly where the overall noise rate is $50\%$, i.e., half of instances are mislabeled.
We ran $10$ trials and reported the means and standard deviations of the test accuracy in \cref{tab:image}.

We can observe that the proposed method generally outperforms the baseline methods.
It is worth noting that estimating a full transition matrix for each instance (Adaptation) may improve the accuracy over CCE when the number of classes is relatively small, but when there are more classes (e.g., CIFAR-100), the performance may drop drastically because it requires an additional $K \times K$ output and the estimation error could be high.
On the contrary, the complexity of our single-parameter approximation does not increase as the number of classes increases.
Additionally, in the ridgeline plots in \cref{fig:ridge}, we can observe the separation of instances with flipped/original labels using the learned confidence $C$, which may explain the performance improvement.


\begin{figure}[t]
\centering
\begin{subfigure}{0.49\textwidth}
\centering
\includegraphics[width=\textwidth]
{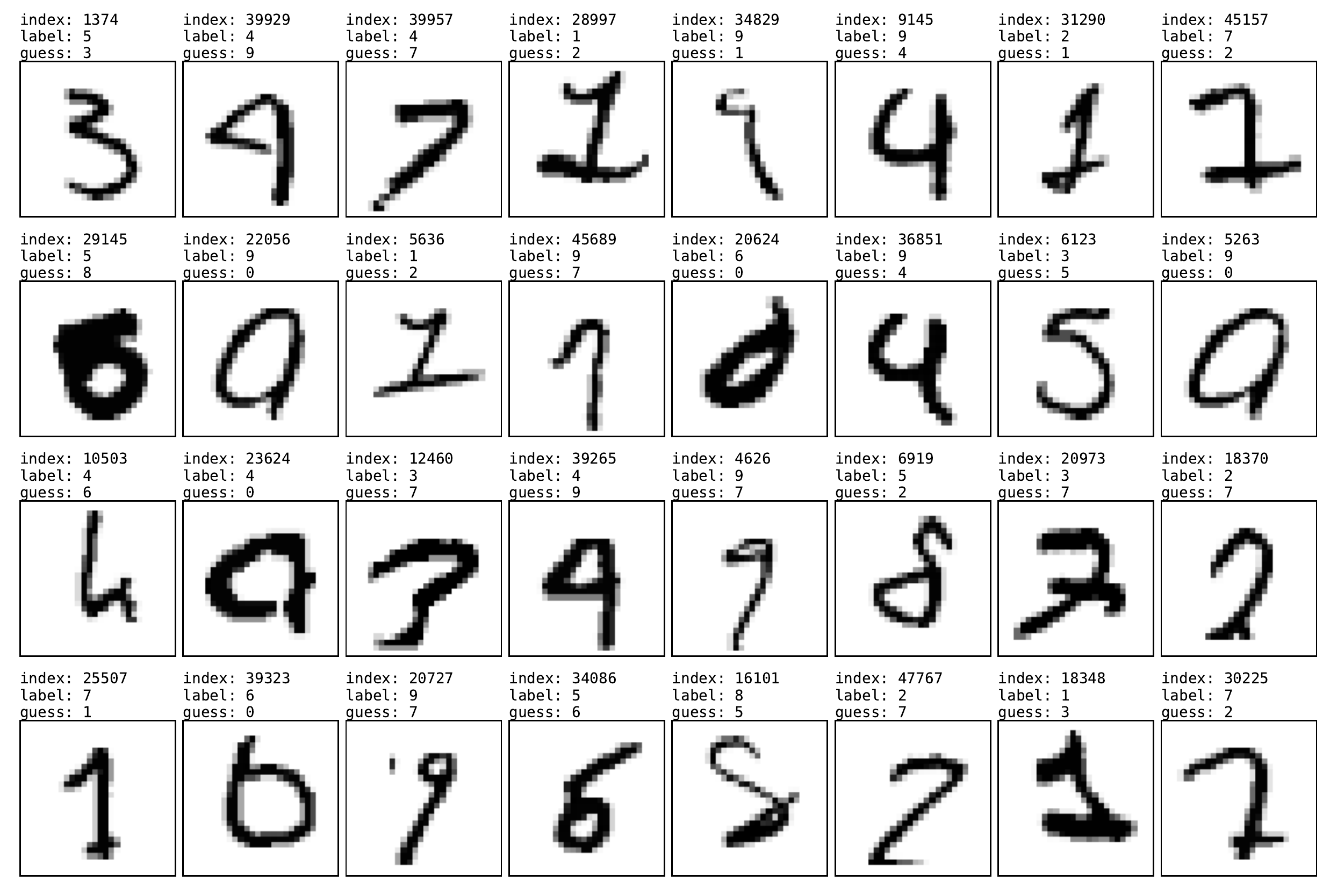}
\caption{MNIST}
\end{subfigure}
\hfill
\begin{subfigure}{0.49\textwidth}
\centering
\includegraphics[width=\textwidth]
{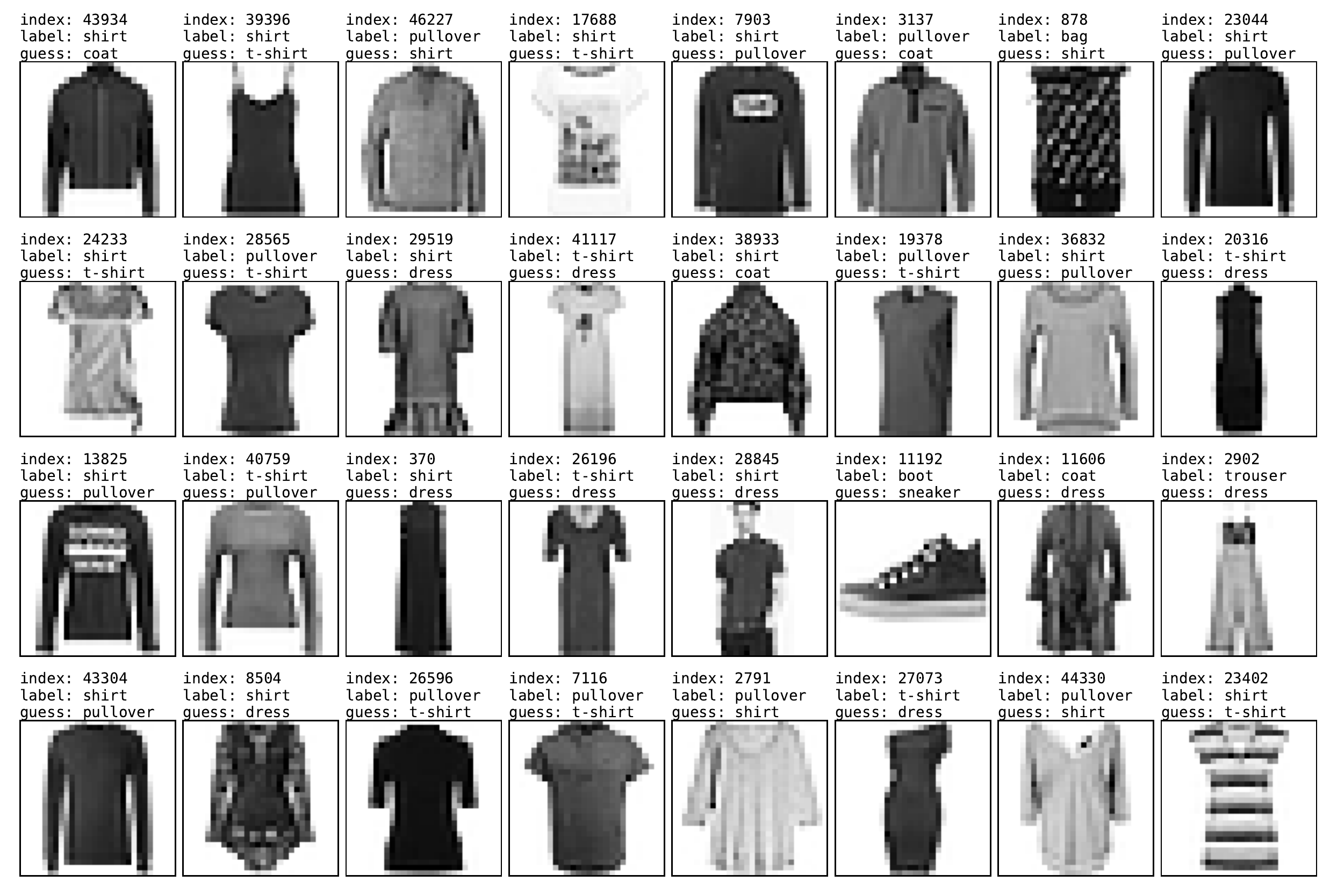}
\caption{FMNIST}
\end{subfigure}

\begin{subfigure}{0.49\textwidth}
\centering
\includegraphics[width=\textwidth]
{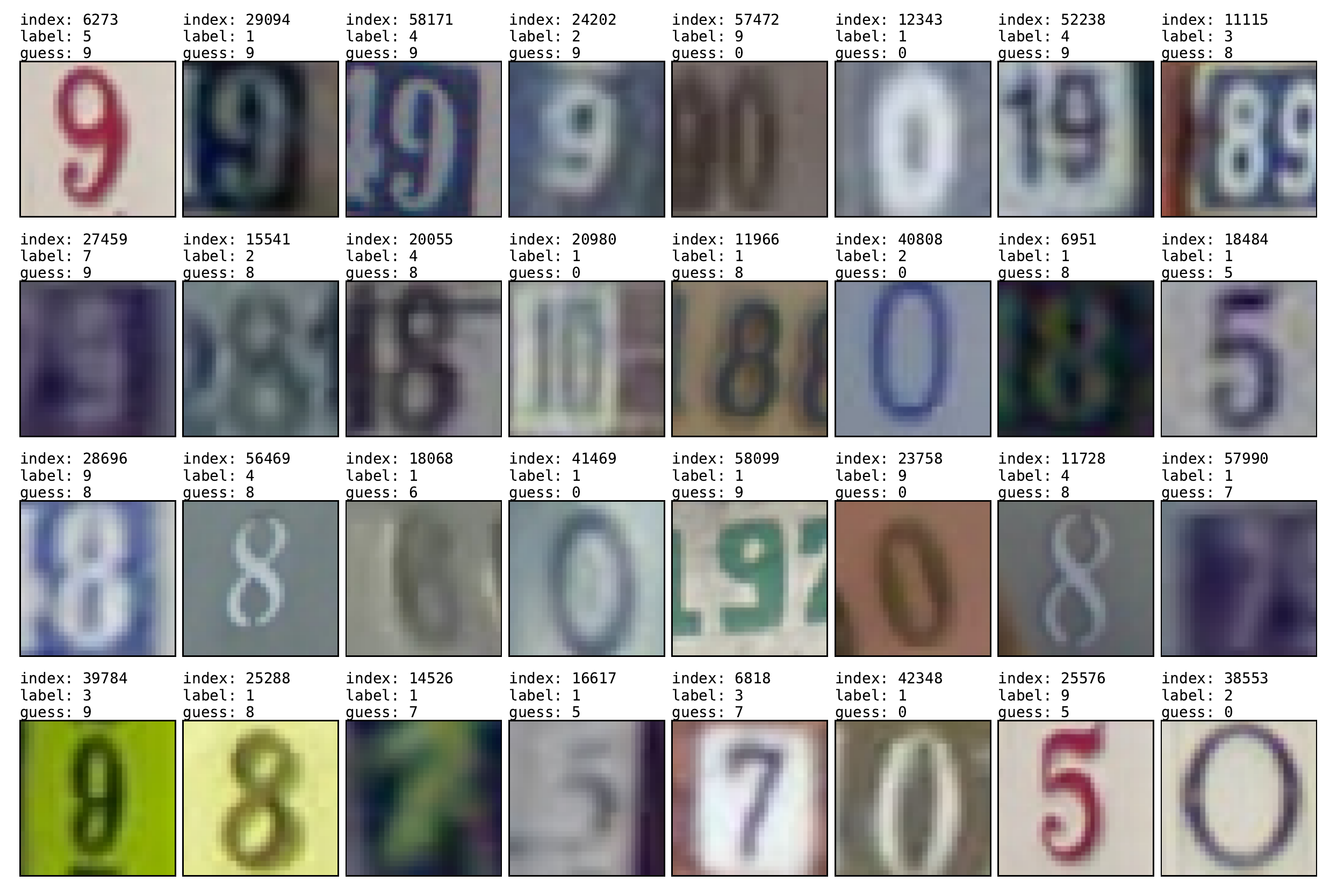}
\caption{SVHN}
\end{subfigure}
\hfill
\begin{subfigure}{0.49\textwidth}
\centering
\includegraphics[width=\textwidth]
{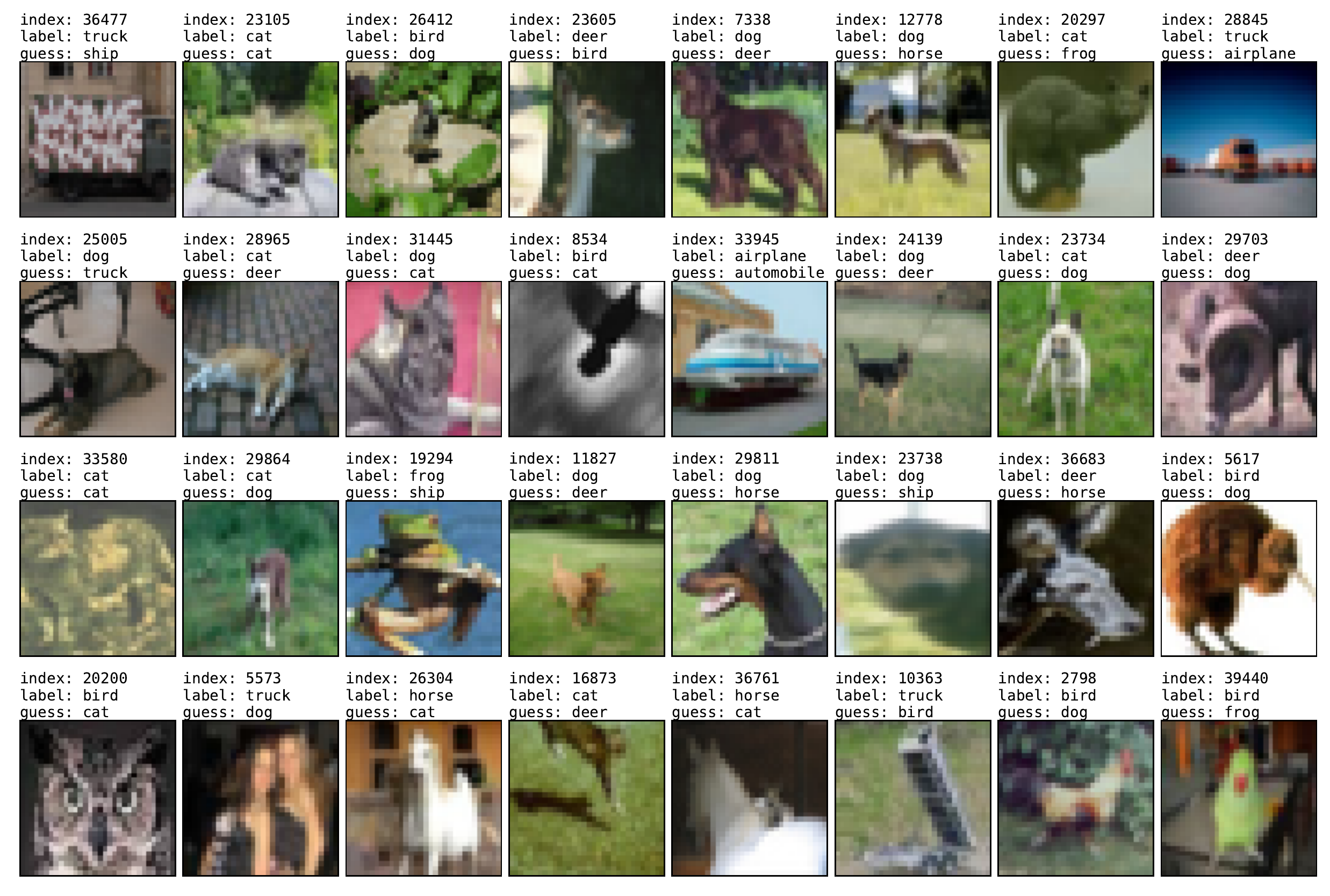}
\caption{CIFAR-10}
\end{subfigure}
\caption{%
The $32$ most \textbf{low-confidence training examples} in the MNIST, FMNIST, SVHN and CIFAR-10 datasets, ordered left-right, top-down by increasing confidence.
The index in the dataset, original label, and predicted label are annotated above each image.
}
\label{fig:low_confidence}
\end{figure}


\paragraph{Detecting ambiguous/mislabeled instances.}

Next, we demonstrate that the proposed method can be used for detecting ambiguous or possibly mislabeled instances.
We trained the model on the original datasets with the proposed method.
A benefit of using a single-parameter approximation is that it naturally derives an order of the training examples.
We sorted the training examples via the confidence and showed the $32$ most low-confidence ones in in \cref{fig:low_confidence}.
Results of other datasets are provided in \cref{app:ex}.

We can observe that, surprisingly, in these supposedly clean datasets, a number of instance might be mislabeled.
In MNIST and SVHN, we found clearly mislabeled images.
There are ambiguous images such as 2-7 and 4-9 pairs in MNIST and shirt/T-shirt/pullover/coat photos in Fashion-MNIST.
In CIFAR-10, it is interesting that images in the animal category are more likely to have a low confidence.
We conjecture that the \emph{spurious correlation} between the object and the background color plays an important role.
We also found multi-modality issues, e.g., kiwi, owl, and chicken are all labeled as bird but are not \emph{visually prototypical} birds.
This phenomenon suggests the possibility of using the proposed method for diagnosing label issues in large-scale datasets.


\begin{table}[t]
\centering
\caption{%
\textbf{Performance} ($\%$) on the GLUE benchmark for natural language understanding.
We reported Matthews correlation coefficient on CoLA, F1 score/accuracy on MRPC and QQP, and accuracy otherwise.
MNLI-(m/mm) denotes MultiNLI matched/mismatched, respectively.
}
\label{tab:glue}
\resizebox{\linewidth}{!}{%
\begin{tabular}{cccccccccc}
\toprule
& CoLA & SST2 & MRPC & QQP & MNLI-(m/mm) & QNLI & RTE & WNLI
\\
\midrule

CCE & $54.76$ & $92.55$ & $88.04$/$82.35$ & $87.80$/$90.96$ & $83.83$/$84.31$ & $90.77$ & $66.43$ & $50.70$ \\
ICE & $57.83$ & $92.20$ & $89.54$/$85.05$ & $87.85$/$90.92$ & $83.81$/$84.36$ & $91.14$ & $63.90$ & $56.34$ \\

\bottomrule
\end{tabular}
} 
\end{table}

\begin{table}[t]
\centering
\caption{%
Selected \textbf{low-confidence training examples} in the CoLA dataset.
}
\label{tab:low_confidence}
\resizebox{\linewidth}{!}{%
\begin{tabular}{cccp{.55\linewidth}}
\toprule
Index & Label & Guess & \multicolumn{1}{c}{Text}
\\

\midrule
390 &
acceptable & unacceptable &
\textcolor{xkcd red}{
\lstinline| He I often sees Mary. |
}
\\

\midrule
7756 &
acceptable & unacceptable &
\textcolor{xkcd red}{
\lstinline| That monkey is ate the banana. |
}
\\

\midrule
8332 &
acceptable & unacceptable &
\textcolor{xkcd red}{
\lstinline| I wanted Jimmy for to come with me. |
}
\\

\midrule
2801 &
unacceptable & acceptable &
\textcolor{xkcd blue}{
\lstinline| Paula hit the sticks. |
}
\\

\midrule
2479 &
unacceptable & acceptable &
\textcolor{xkcd blue}{
\lstinline| Kelly buttered the bread with butter. |
}
\\

\midrule
6795 &
unacceptable & acceptable &
\textcolor{xkcd blue}{
\lstinline| Henry wanted to possibly marry Fanny. |
}
\\

\bottomrule
\end{tabular}
} 
\end{table}


\subsection{Text classification}
\label{ssec:text}

We discovered that noisy label issues also exist in text datasets.
We conducted similar experiments on the GLUE benchmark \citep{wang2018glue}, which is a collection of datasets for natural language understanding.
We trained a BERT-base model pretrained using a masked language modeling (MLM) objective \citep{devlin2018bert} with a default AdamW optimizer \citep{loshchilov2017decoupled}.
The performance in terms of the suggested evaluation metric was reported in \cref{tab:glue}.

We can observe that, except on the RTE dataset, the performance was improved or approximately the same compared with the default CCE method, which shows the benefits of using instance-specifically adjusted confidences.
Although, if the dataset is relatively clean, the improvement might be marginal.

We also found mislabeled or ambiguous instances in these datasets.
A typical example is the Corpus of Linguistic Acceptability (CoLA) \citep{warstadt2019neural} dataset, which consists of English grammatical acceptability judgments.
Six selected low-confidence training examples are given in \cref{tab:low_confidence}. 
We found that several ungrammatical sentences were mislabeled as acceptable, and some \emph{syntactically acceptable} sentences were labeled as unacceptable by annotators possibly because they have \emph{semantic errors}.
In this way, we may use the proposed method to probe if the model prediction is consistent with our intent.
More results are provided in \cref{app:ex}.

\section{Conclusion}
We have introduced a novel variational approximation of the instance-dependent noise (IDN) model, referred to as \emph{instance-confidence embedding} (ICE).
Compared with existing methods based on the class-conditional noise (CCN) assumption, the proposed method is able to capture instance-specific noise information and consequently improve the classification performance.
The use of the one-dimensional instance embedding naturally derives an order of training examples which can be used for detecting ambiguous or mislabeled instances.
For future directions, it is interesting to explore its combination with other training techniques and its extensions in data cleansing, learning with rejection, or active learning.

\clearpage
\section*{Acknowledgement}
We thank Gang Niu, Xi Huang, Nontawat Charoenphakdee, Han Bao, Masato Ishii, Yoshihiro Nagano, and Shota Nakajima for helpful discussion.
We also would like to thank the Supercomputing Division, Information Technology Center, the University of Tokyo, for providing the Reedbush supercomputer system.
YZ was supported by Microsoft Research Asia D-CORE program and Junior Research Associate (JRA) program at RIKEN.
MS was supported by JST AIP Acceleration Research Grant Number JPMJCR20U3 and the Institute for AI and Beyond, UTokyo, Japan.

\bibliography{references}


\clearpage
\appendix
\everymath{\displaystyle}

\newgeometry{
  margin=1.5cm,
  includefoot,
  footskip=30pt,
}


\clearpage
\section{Gradient analysis}
\label{app:gradient}

\begin{figure}
\centering
\includegraphics[width=\linewidth]
{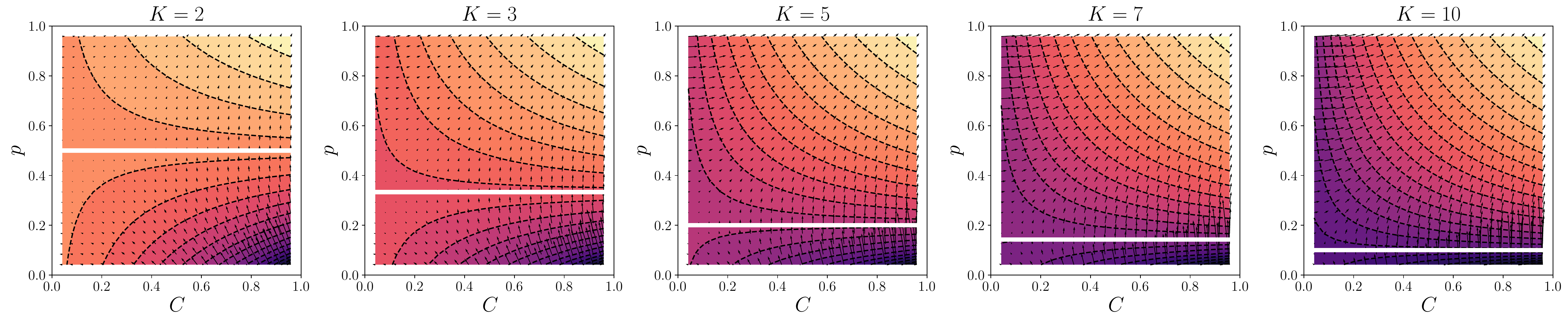}
\caption{%
Contours of the log-likelihood w.r.t. $\vp_i$ and $C$ 
using the linear interpolation (\cref{eq:linear}) 
for $K$ in $\{2, 3, 5, 7, 10\}$.
}
\label{fig:gradient_pc}
\end{figure}

\begin{figure}
\centering
\includegraphics[width=\linewidth]
{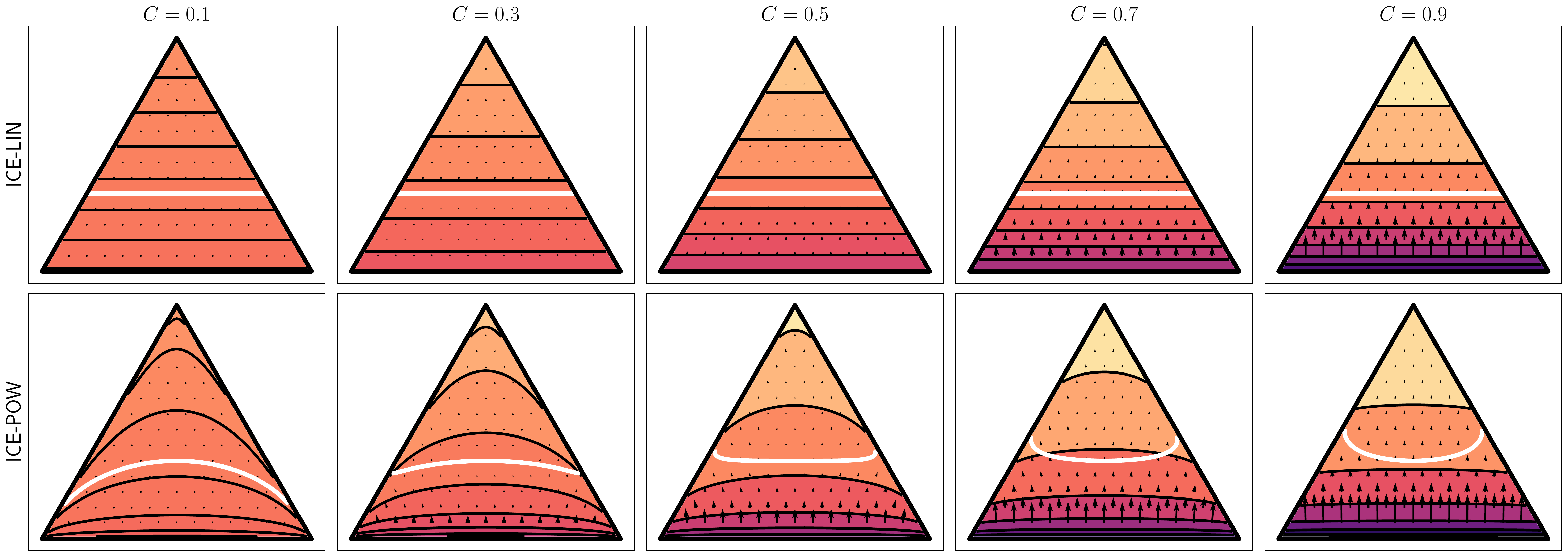}
\caption{%
Contours of the log-likelihood on the simplex 
when $K = 3$ 
using the linear interpolation (\cref{eq:linear}, top) 
and the power transformation (\cref{eq:power}, bottom) 
for $C$ in $\{0.1, 0.3, 0.5, 0.7, 0.9\}$.
}
\label{fig:gradient_simplex}
\end{figure}

In this section, we provide a basic gradient analysis and visualization for our proposed method.

The gradients of the log-likelihood using \cref{eq:linear,eq:power} are
\begin{align}
\label{eq:gradient_linear}
\phantom{\text{(power transformation)}}
  \diffp{}{C} \log(\vq_i)
&= 
  \frac{\vp_i - \frac1K}{C \vp_i + (1 - C) \frac1K}
,
& \text{(linear interpolation)}
\\
\label{eq:gradient_power}
  \diffp{}{C} \log(\vq_i)
&=
  \sum_{j=1}^K \vp_j^C \log \frac{\vp_i}{ \vp_j}
,
& \text{(power transformation)}
\end{align}
respectively.
Their sign boundaries are $\vp_i = \frac1K$ and $\vp_i = e^{-\Eta(\vq, \vp)}$, respectively, where $\Eta(\cdot, \cdot)$ denotes the cross-entropy.
We can find that for the linear interpolation (\cref{eq:linear}), when $\vp_\yt < \frac1K$, $\diffp{L}{C} < 0$ and when $\vp_\yt > \frac1K$, $\diffp{L}{C} > 0$.
Similarly, for the power transformation (\cref{eq:power}), when $\vp_\yt < e^{-\Eta(\vq, \vp)}$, $\diffp{L}{C} < 0$ and when $\vp_\yt > e^{-\Eta(\vq, \vp)}$, $\diffp{L}{C} > 0$.

The contours of the likelihood for different parameters are plotted in \cref{fig:gradient_pc,fig:gradient_simplex}.
Note that the class-posterior $\vp$ is obtained from a neural network, so it can be influenced by other instances, especially adjacent instances.
On the other hand, the confidence $C$ is obtained via instance embedding, so it can take any value independently.
If the predicted class-posterior $\vp_\yt$ for an instance $x$ is low (e.g., because this instance is mislabeled and the majority of adjacent instances are predicted to belong to other classes), then the classifier tends to decrease its confidence value so as not to overfit this possibly mislabeled instance.
The gradient magnitude is the largest when $C$ is high and $\vp_\yt$ is low (confident wrong prediction), the smallest when both $C$ and $\vp_\yt$ are high (confident correct prediction), and in the middle when $C$ is low (uncertain prediction like a random guess).
In this way, we can equip the neural network model with an option of changing the confidence of prediction for individual training examples to mitigate overfitting possibly mislabeled instances.


\clearpage
\section{Linear interpolation}
\label{app:lerp}

\begin{figure}
\centering
\includegraphics[width=.6\linewidth]
{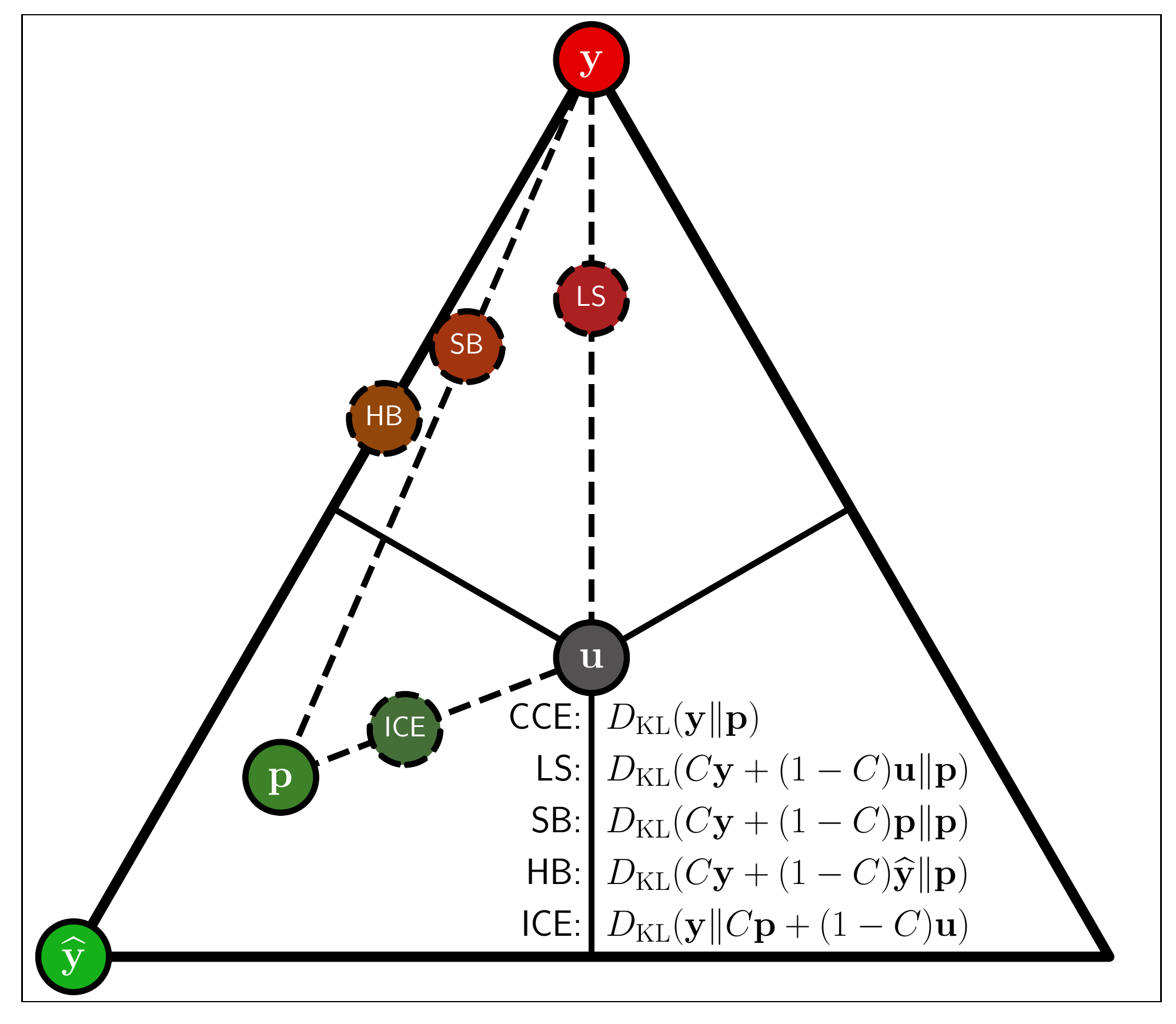}
\caption{%
Illustration of related methods, including the categorical cross-entropy (CCE), label smoothing (LS), soft/hard bootstrapping loss (SB/HB), and the proposed instance-confidence embedding (ICE) with the linear transformation (\cref{eq:linear}).
}
\label{fig:smoothing}
\end{figure}

Linear interpolation between some properties of an instance and other value is a widely used technique for regularization in machine learning, such as the bootstrapping loss \cite{reed2015training} and the label smoothing technique \citep{szegedy2016rethinking, pereyra2017regularizing, lukasik2020does}.
In this section, we briefly summarize related techniques and compare their differences.

Concretely, let $\vp$ be the predicted probability vector for $Y$ (\cref{eq:classifier}), $\vy$ be the one-hot vector for the observed label, $\widehat{\vy} = \argmax \vp$ is the one-hot vector for the predicted label, $\vu$ be the uniform probability vector ($\vu_i = \frac1K$ for $i \in \{1, \dots, K\}$).
Here, $\vp, \vy, \widehat{\vy}, \vu \in \Delta^{K-1}$ are all in the probability simplex.
Let $C \in [0, 1]$ be a scalar linear interpolation parameter.

Then, as also shown in \cref{fig:smoothing}, the learning objectives are equivalent to the following KL-divergences:
\begin{align}
\phantom{\text{(instance-confidence embedding)}}
  D_\KL\diver{\vy}{\vp},
&& \text{(categorical cross-entropy)}
\\
  D_\KL\diver{C\vy + (1 - C)\vu}{\vp},
&& \text{(label smoothing)}
\\
  D_\KL\diver{C\vy + (1 - C)\vp}{\vp},
&& \text{(soft bootstrapping loss)}
\\
  D_\KL\diver{C\vy + (1 - C)\widehat{\vy}}{\vp},
&& \text{(hard bootstrapping loss)}
\\
  D_\KL\diver{\vy}{C\vp + (1 - C)\vu}.
&& \text{(instance-confidence embedding)}
\end{align}
We can see that the label smoothing and the bootstrapping loss methods smooth the target, but the proposed ICE method smooths the prediction.
Note that it is impossible to let $C$ be an instance-dependent parameter in other methods, because when $C = 0$, the supervision signal $Y$ can be completely lost.

Another technique using linear interpolation is \emph{mixup} \citep{zhang2018mixup}, which also interpolates the input features $X$ between two instances.
Therefore its characteristics could be more different than the methods mentioned above.


\clearpage
\section{Experiments}
\label{app:ex}


\begin{figure}[t]
\centering
\includegraphics[width=\textwidth]
{ridge_lin.pdf}
\includegraphics[width=\textwidth]
{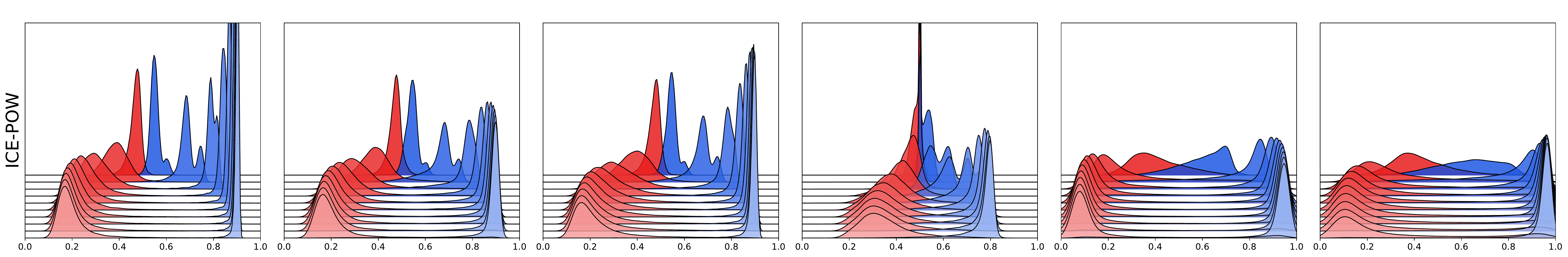}
\caption{%
\textbf{Ridgeline plots} of the confidence $C$ during training.
The density is estimated via Gaussian kernel density estimation (KDE).
The red/blue curves represent the confidence of instances with flipped/original labels, respectively.
}
\label{fig:ridge_full}
\end{figure}


\subsection{Image classification}

\paragraph{Data.}
We used the 
MNIST,\footnote{%
MNIST \citep{lecun1998gradient}
\url{http://yann.lecun.com/exdb/mnist/}
}
Fashion-MNIST,\footnote{%
Fashion-MNIST \citep{xiao2017fashion}
\url{https://github.com/zalandoresearch/fashion-mnist}
}
Kuzushiji-MNIST,\footnote{%
Kuzushiji-MNIST \citep{clanuwat2018deep}
\url{http://codh.rois.ac.jp/kmnist/}
}
SVHN,\footnote{%
SVHN \citep{netzer2011reading}
\url{http://ufldl.stanford.edu/housenumbers}
}
CIFAR-10, and CIFAR-100\footnote{%
CIFAR-10, CIFAR-100 \citep{krizhevsky2009learning}
\url{https://www.cs.toronto.edu/~kriz/cifar.html}
}
datasets.
The MNIST, Fashion-MNIST, Kuzushiji-MNIST datasets contain $28 \times 28$ grayscale images in $10$ classes.
The size of the training set is $60000$ and the size of the test set is $10000$.
The SVHN dataset contains $32 \times 32$ colour images in $10$ classes.
The size of the training set is $73257$ and the size of the test set is $26032$.
The CIFAR-10 and CIFAR-100 datasets contain $32 \times 32$ colour images in $10$ classes and in $100$ classes, respectively.
The size of the training set is $50000$ and the size of the test set is $10000$.
We used $20\%$ of the training sets for validation.
We added synthetic label noise into the training and validation sets.
The test sets were not modified.

\paragraph{Models.}

For MNIST, Fashion-MNIST, and Kuzushiji-MNIST, we used a sequential convolutional neural network with the following structure:
\texttt{Conv2d}(channel=$32$) $\times2$,
\texttt{Conv2d}(channel=$64$) $\times2$,
\texttt{MaxPool2d}(size=$2$),
\texttt{Linear}(dim=$128$),
\texttt{Dropout}(p=$0.5$),
\texttt{Linear}(dim=$10$).
The kernel size of convolutional layers is $3$, and rectified linear unit (ReLU) is applied after the convolutional layers and linear layers except the last one.
For SVHN, CIFAR-10 and CIFAR-100, we used a ResNet-18 model \citep{he2016deep}.
To ensure that $C \in [0, 1]$, we simply apply the sigmoid function that maps $\R$ to $[0, 1]$ to the embedding.

\paragraph{Optimization.}

For MNIST, Fashion-MNIST, and Kuzushiji-MNIST, we used an Adam optimizer \citep{kingma2014adam} with batch size of $512$ and learning rate of $\num{1e-3}$.
The model was trained for $2000$ iterations ($17.07$ epochs) and the learning rate decayed exponentially to $\num{1e-4}$.
For CIFAR-10 and CIFAR-100, we used a stochastic gradient descent (SGD) optimizer with batch size of $512$, momentum of $0.9$, and weight decay of $\num{1e-4}$.
The learning rate increased from $0$ to $0.1$ linearly for $400$ iterations and decreased to $0$ linearly for $3600$ iterations ($4000$ iterations/$40.96$ epochs in total).
For SVHN, the setting was the same as CIFAR-10 except the model was trained for $1000$ iterations.

\paragraph{Results.}

\begin{figure}[t]
\centering

\begin{subfigure}{0.49\textwidth}
\centering
\includegraphics[width=\textwidth]
{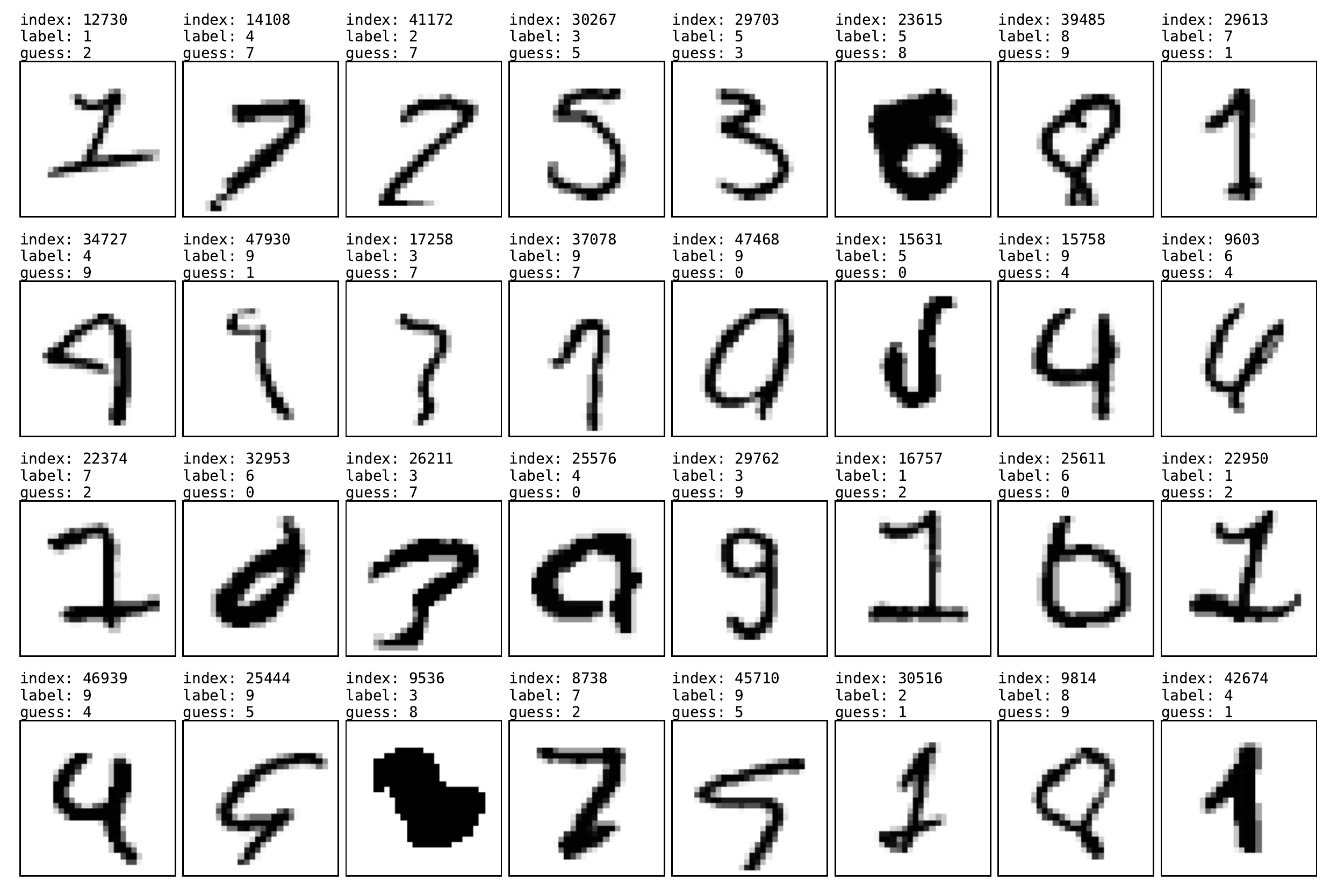}
\caption{MNIST (ICE-LIN)}
\end{subfigure}
\hfill
\begin{subfigure}{0.49\textwidth}
\centering
\includegraphics[width=\textwidth]
{image/image_mnist_pow.pdf}
\caption{MNIST (ICE-POW)}
\end{subfigure}

\begin{subfigure}{0.49\textwidth}
\centering
\includegraphics[width=\textwidth]
{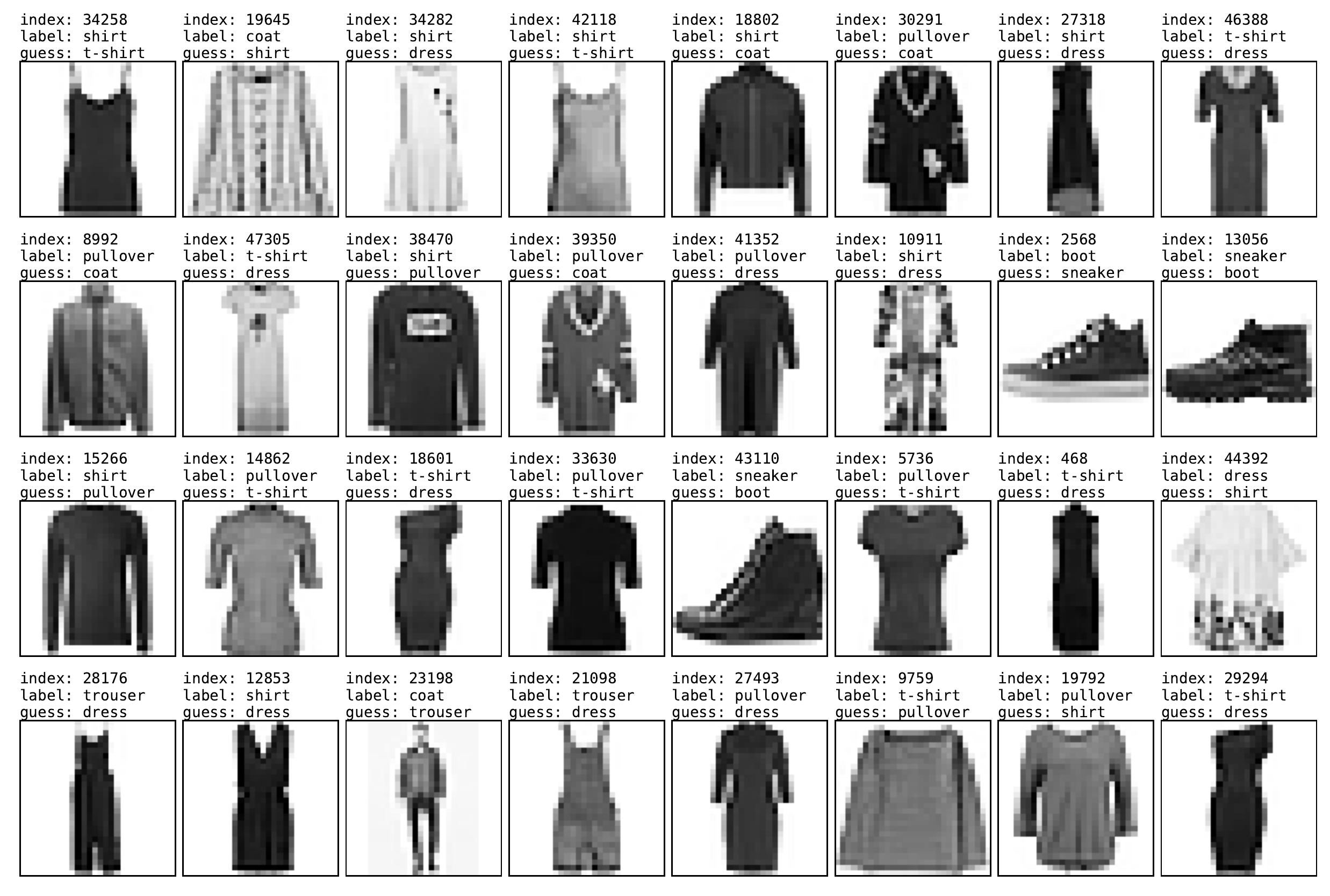}
\caption{FMNIST (ICE-LIN)}
\end{subfigure}
\hfill
\begin{subfigure}{0.49\textwidth}
\centering
\includegraphics[width=\textwidth]
{image/image_fashion-mnist_pow.pdf}
\caption{FMNIST (ICE-POW)}
\end{subfigure}

\begin{subfigure}{0.49\textwidth}
\centering
\includegraphics[width=\textwidth]
{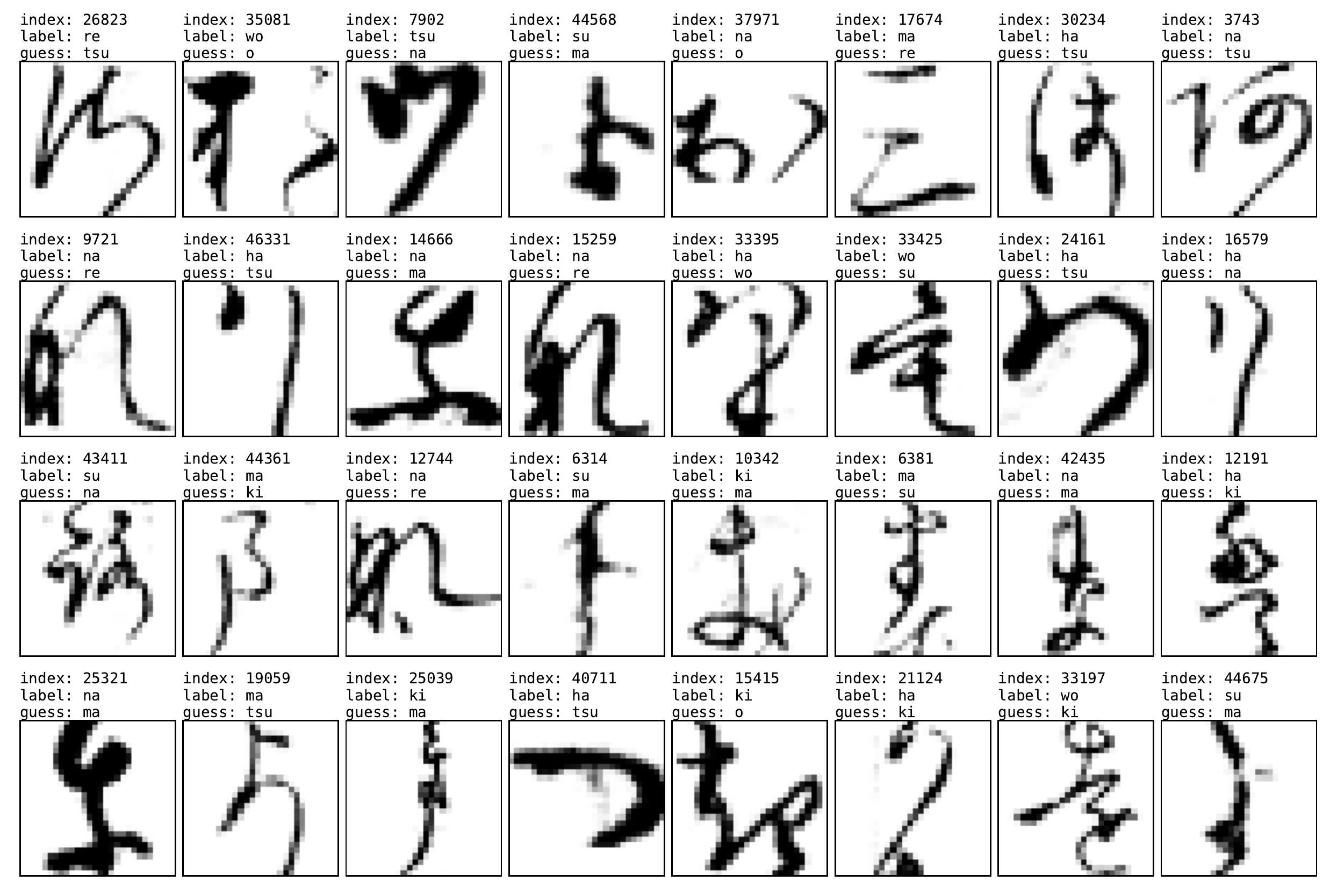}
\caption{KMNIST (ICE-LIN)}
\end{subfigure}
\hfill
\begin{subfigure}{0.49\textwidth}
\centering
\includegraphics[width=\textwidth]
{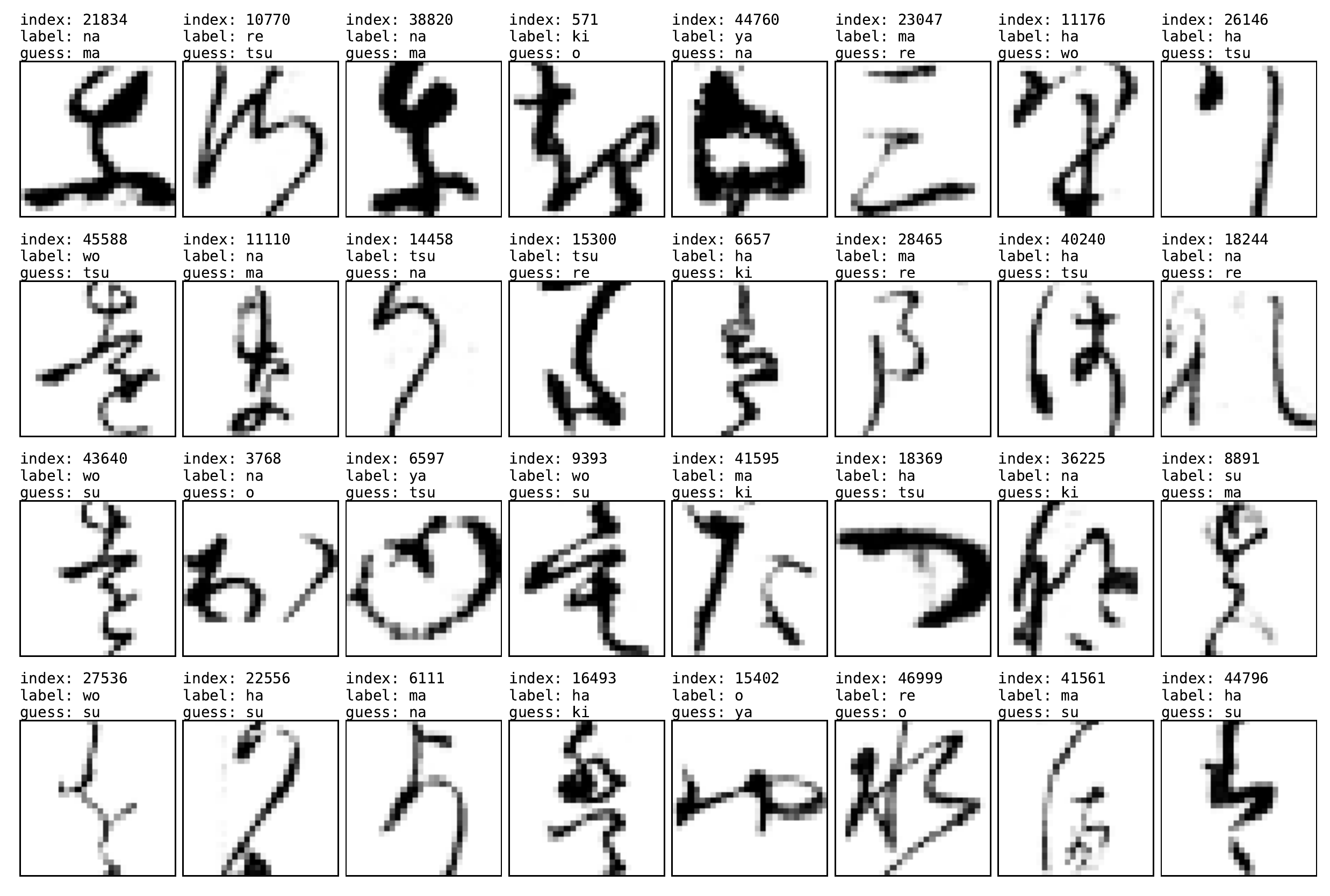}
\caption{KMNIST (ICE-POW)}
\end{subfigure}

\caption{%
The $32$ most \textbf{low-confidence training examples} in the MNIST, FMNIST, and KMNIST datasets, ordered left-right, top-down by increasing confidence.
}
\label{fig:mnist}
\end{figure}

\begin{figure}[t]
\centering

\begin{subfigure}{0.49\textwidth}
\centering
\includegraphics[width=\textwidth]
{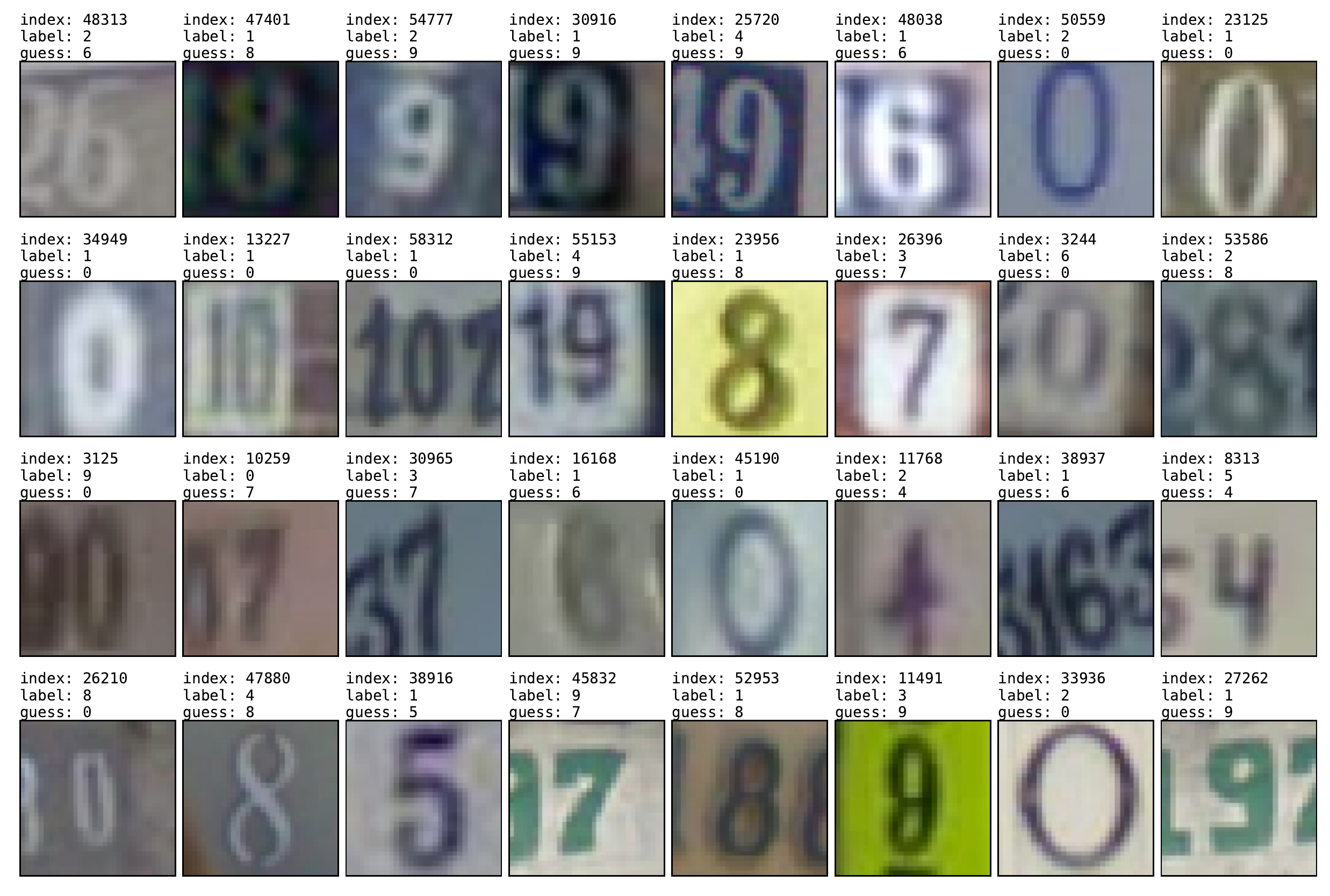}
\caption{SVHN (ICE-LIN)}
\end{subfigure}
\hfill
\begin{subfigure}{0.49\textwidth}
\centering
\includegraphics[width=\textwidth]
{image/image_svhn_pow.pdf}
\caption{SVHN (ICE-POW)}
\end{subfigure}

\begin{subfigure}{0.49\textwidth}
\centering
\includegraphics[width=\textwidth]
{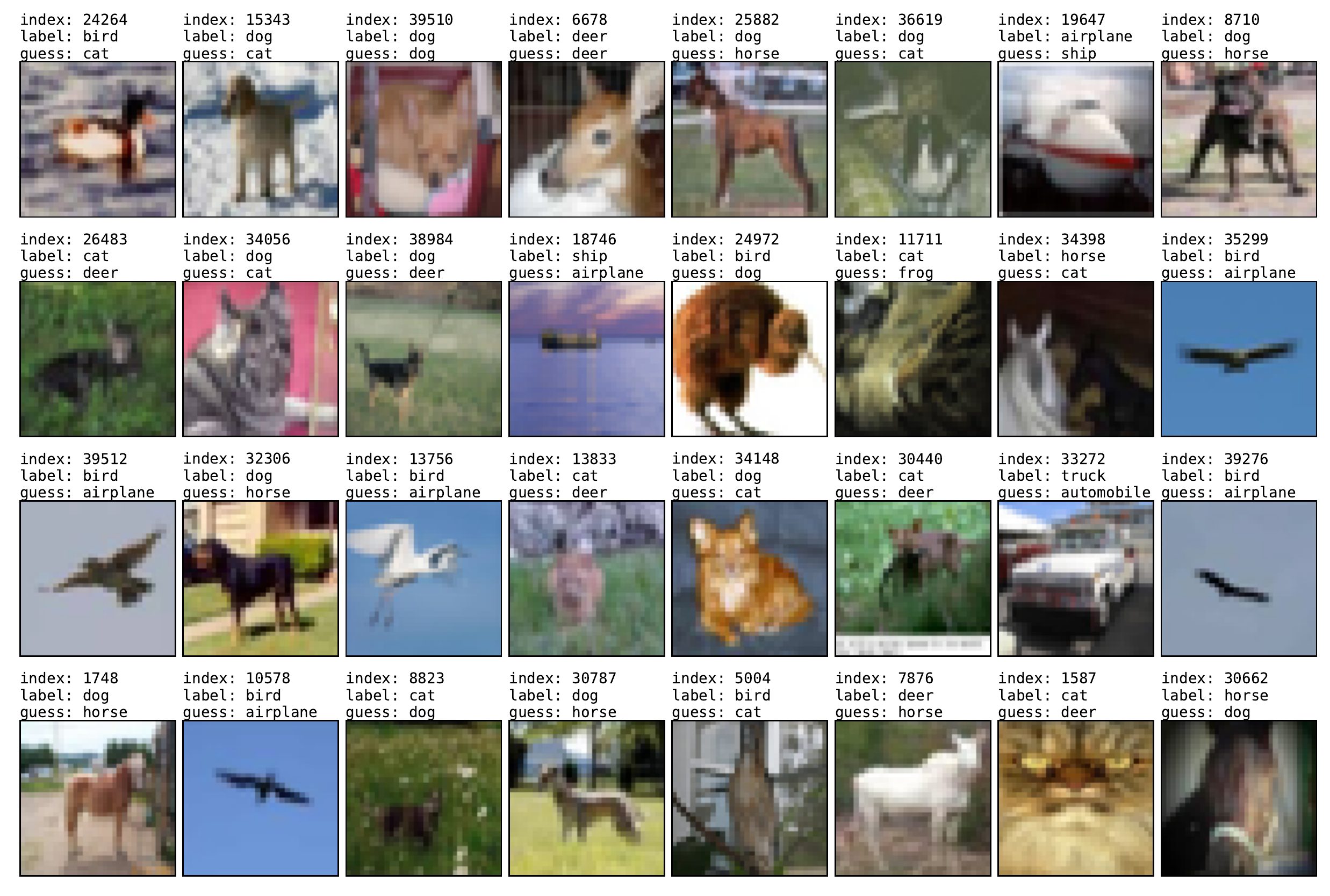}
\caption{CIFAR-10 (ICE-LIN)}
\end{subfigure}
\hfill
\begin{subfigure}{0.49\textwidth}
\centering
\includegraphics[width=\textwidth]
{image/image_cifar10_pow.pdf}
\caption{CIFAR-10 (ICE-POW)}
\end{subfigure}

\begin{subfigure}{0.49\textwidth}
\centering
\includegraphics[width=\textwidth]
{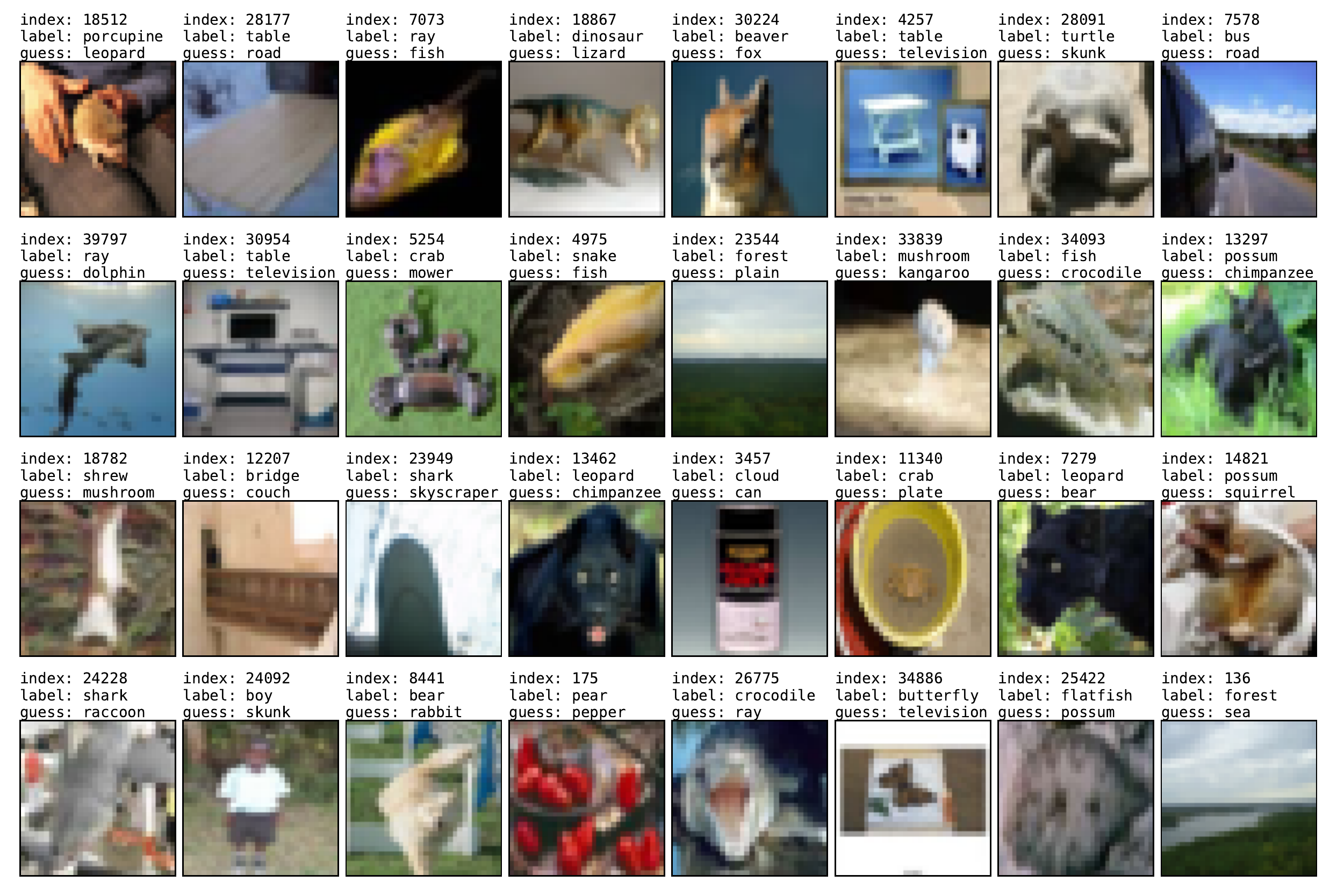}
\caption{CIFAR-100 (ICE-LIN)}
\end{subfigure}
\hfill
\begin{subfigure}{0.49\textwidth}
\centering
\includegraphics[width=\textwidth]
{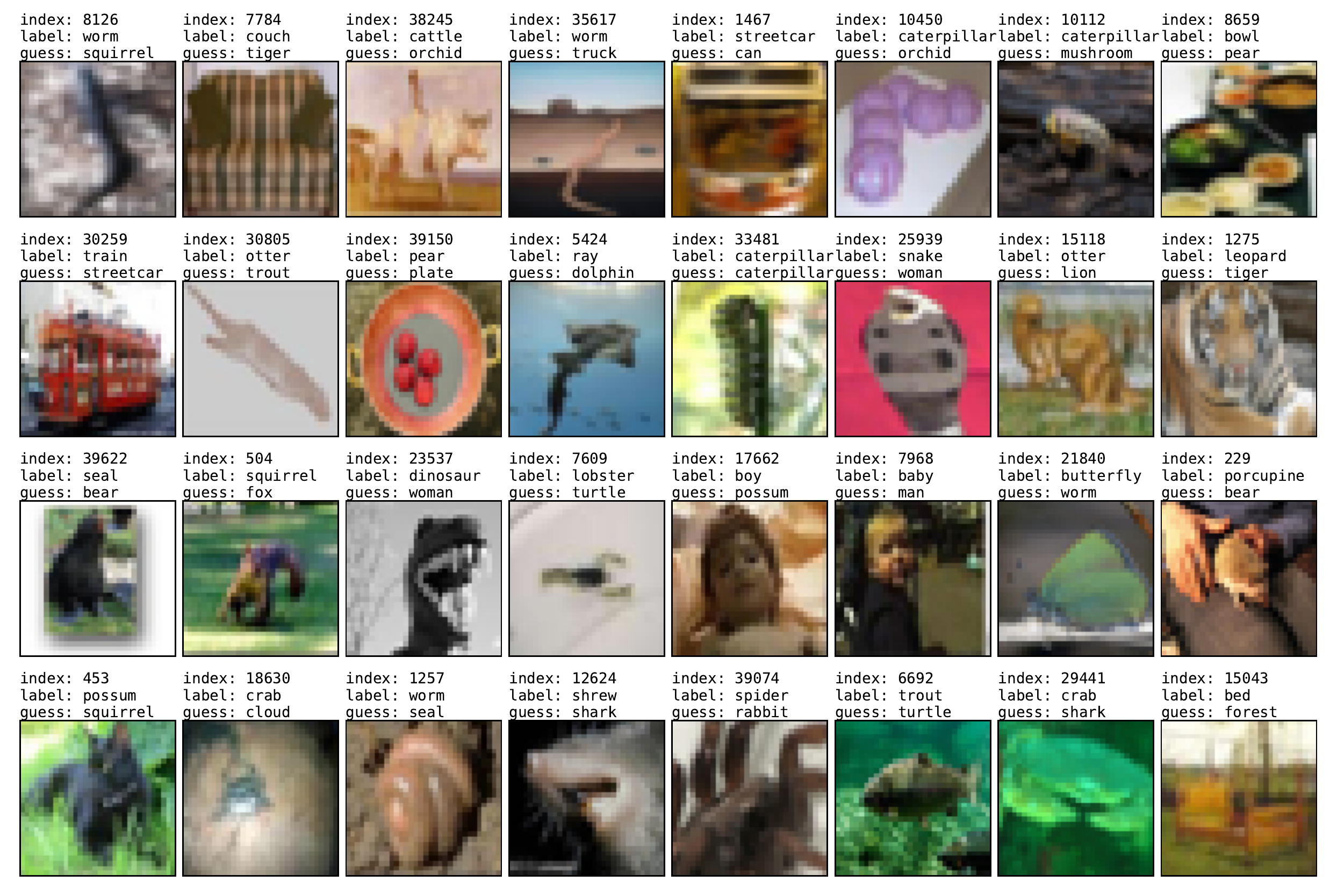}
\caption{CIFAR-100 (ICE-POW)}
\end{subfigure}

\caption{%
The $32$ most \textbf{low-confidence training examples} in the SVHN, CIFAR-10 and CIFAR-100 datasets, ordered left-right, top-down by increasing confidence.
}
\label{fig:cifar}
\end{figure}

The ridgeline plots of the confidence $C$ during training are given in \cref{fig:ridge_full}.
Low-confidence training examples are given in \cref{fig:mnist,fig:cifar}, which are partially presented in \cref{fig:low_confidence}.


\subsection{Text classification}

We implemented the BERT-base model \citep{devlin2018bert} using PyTorch \citep{paszke2019pytorch} and HuggingFace's \texttt{transformers} \citep{wolf2019huggingface} libraries.
We used a pretrained model%
\footnote{%
\texttt{bert-base-cased}: 
\url{https://huggingface.co/bert-base-cased}
}
and an AdamW optimizer \citep{loshchilov2017decoupled}.
The batch size was $32$ and the weight decay was $0.01$, otherwise we used the default hyperparameters.
The model was trained on $4$ NVIDIA Tesla P100 GPUs in parallel with the mixed precision training option (\texttt{fp16}) enabled.
For the CoLA, MRPC, RTE, and WNLI datasets, the model was trained for $5$ epochs and otherwise $3$ epochs.
Low-confidence training examples are given in \cref{tab:cola,tab:sst2,tab:mrpc,tab:qqp,tab:mnli,tab:qnli,tab:rte,tab:wnli}.


\clearpage


\clearpage


\clearpage
\begin{table}
\centering
\caption{The Corpus of Linguistic Acceptability (CoLA)}
\label{tab:cola}
\begin{tabular}{cp{0.9\linewidth}}
\toprule
Index & Text
\\


\midrule
390 &
\begin{tabular}[c]{@{}l@{}}
\underline{\texttt{%
\textcolor{xkcd red}{label}:acceptable
\textcolor{xkcd blue}{guess}:unacceptable}}
\vspace{.2em}\\
\begin{minipage}{\linewidth}\begin{flushleft}
\texttt{\textcolor{xkcd grey}{sentence}:}\lstinline| He I often sees Mary. |
\end{flushleft}\end{minipage}
\end{tabular}
\\


\midrule
5766 &
\begin{tabular}[c]{@{}l@{}}
\underline{\texttt{%
\textcolor{xkcd red}{label}:acceptable
\textcolor{xkcd blue}{guess}:unacceptable}}
\vspace{.2em}\\
\begin{minipage}{\linewidth}\begin{flushleft}
\texttt{\textcolor{xkcd grey}{sentence}:}\lstinline| Heidi believes any description of herself. |
\end{flushleft}\end{minipage}
\end{tabular}
\\


\midrule
2801 &
\begin{tabular}[c]{@{}l@{}}
\underline{\texttt{%
\textcolor{xkcd red}{label}:unacceptable
\textcolor{xkcd blue}{guess}:acceptable}}
\vspace{.2em}\\
\begin{minipage}{\linewidth}\begin{flushleft}
\texttt{\textcolor{xkcd grey}{sentence}:}\lstinline| Paula hit the sticks. |
\end{flushleft}\end{minipage}
\end{tabular}
\\


\midrule
1522 &
\begin{tabular}[c]{@{}l@{}}
\underline{\texttt{%
\textcolor{xkcd red}{label}:unacceptable
\textcolor{xkcd blue}{guess}:acceptable}}
\vspace{.2em}\\
\begin{minipage}{\linewidth}\begin{flushleft}
\texttt{\textcolor{xkcd grey}{sentence}:}\lstinline| That the sun is out was obvious. |
\end{flushleft}\end{minipage}
\end{tabular}
\\


\midrule
8332 &
\begin{tabular}[c]{@{}l@{}}
\underline{\texttt{%
\textcolor{xkcd red}{label}:acceptable
\textcolor{xkcd blue}{guess}:unacceptable}}
\vspace{.2em}\\
\begin{minipage}{\linewidth}\begin{flushleft}
\texttt{\textcolor{xkcd grey}{sentence}:}\lstinline| I wanted Jimmy for to come with me. |
\end{flushleft}\end{minipage}
\end{tabular}
\\


\midrule
300 &
\begin{tabular}[c]{@{}l@{}}
\underline{\texttt{%
\textcolor{xkcd red}{label}:acceptable
\textcolor{xkcd blue}{guess}:unacceptable}}
\vspace{.2em}\\
\begin{minipage}{\linewidth}\begin{flushleft}
\texttt{\textcolor{xkcd grey}{sentence}:}\lstinline| They failed to tell me which problem the sooner I solve, the quicker the folks up at corporate headquarters. |
\end{flushleft}\end{minipage}
\end{tabular}
\\


\midrule
7813 &
\begin{tabular}[c]{@{}l@{}}
\underline{\texttt{%
\textcolor{xkcd red}{label}:acceptable
\textcolor{xkcd blue}{guess}:unacceptable}}
\vspace{.2em}\\
\begin{minipage}{\linewidth}\begin{flushleft}
\texttt{\textcolor{xkcd grey}{sentence}:}\lstinline| I went to the shop for to get bread. |
\end{flushleft}\end{minipage}
\end{tabular}
\\


\midrule
5904 &
\begin{tabular}[c]{@{}l@{}}
\underline{\texttt{%
\textcolor{xkcd red}{label}:acceptable
\textcolor{xkcd blue}{guess}:unacceptable}}
\vspace{.2em}\\
\begin{minipage}{\linewidth}\begin{flushleft}
\texttt{\textcolor{xkcd grey}{sentence}:}\lstinline| It hailed. |
\end{flushleft}\end{minipage}
\end{tabular}
\\


\midrule
4159 &
\begin{tabular}[c]{@{}l@{}}
\underline{\texttt{%
\textcolor{xkcd red}{label}:unacceptable
\textcolor{xkcd blue}{guess}:acceptable}}
\vspace{.2em}\\
\begin{minipage}{\linewidth}\begin{flushleft}
\texttt{\textcolor{xkcd grey}{sentence}:}\lstinline| Fifteen years represent a long period of his life. |
\end{flushleft}\end{minipage}
\end{tabular}
\\


\midrule
2479 &
\begin{tabular}[c]{@{}l@{}}
\underline{\texttt{%
\textcolor{xkcd red}{label}:unacceptable
\textcolor{xkcd blue}{guess}:acceptable}}
\vspace{.2em}\\
\begin{minipage}{\linewidth}\begin{flushleft}
\texttt{\textcolor{xkcd grey}{sentence}:}\lstinline| Kelly buttered the bread with butter. |
\end{flushleft}\end{minipage}
\end{tabular}
\\


\midrule
3846 &
\begin{tabular}[c]{@{}l@{}}
\underline{\texttt{%
\textcolor{xkcd red}{label}:acceptable
\textcolor{xkcd blue}{guess}:acceptable}}
\vspace{.2em}\\
\begin{minipage}{\linewidth}\begin{flushleft}
\texttt{\textcolor{xkcd grey}{sentence}:}\lstinline| They parted the best of friends. |
\end{flushleft}\end{minipage}
\end{tabular}
\\


\midrule
7371 &
\begin{tabular}[c]{@{}l@{}}
\underline{\texttt{%
\textcolor{xkcd red}{label}:unacceptable
\textcolor{xkcd blue}{guess}:acceptable}}
\vspace{.2em}\\
\begin{minipage}{\linewidth}\begin{flushleft}
\texttt{\textcolor{xkcd grey}{sentence}:}\lstinline| The hiker will reach the top of the mountain for an hour. |
\end{flushleft}\end{minipage}
\end{tabular}
\\


\midrule
430 &
\begin{tabular}[c]{@{}l@{}}
\underline{\texttt{%
\textcolor{xkcd red}{label}:unacceptable
\textcolor{xkcd blue}{guess}:acceptable}}
\vspace{.2em}\\
\begin{minipage}{\linewidth}\begin{flushleft}
\texttt{\textcolor{xkcd grey}{sentence}:}\lstinline| It's probable in general that he understands what's going on. |
\end{flushleft}\end{minipage}
\end{tabular}
\\


\midrule
6795 &
\begin{tabular}[c]{@{}l@{}}
\underline{\texttt{%
\textcolor{xkcd red}{label}:unacceptable
\textcolor{xkcd blue}{guess}:acceptable}}
\vspace{.2em}\\
\begin{minipage}{\linewidth}\begin{flushleft}
\texttt{\textcolor{xkcd grey}{sentence}:}\lstinline| Henry wanted to possibly marry Fanny. |
\end{flushleft}\end{minipage}
\end{tabular}
\\


\midrule
1115 &
\begin{tabular}[c]{@{}l@{}}
\underline{\texttt{%
\textcolor{xkcd red}{label}:acceptable
\textcolor{xkcd blue}{guess}:acceptable}}
\vspace{.2em}\\
\begin{minipage}{\linewidth}\begin{flushleft}
\texttt{\textcolor{xkcd grey}{sentence}:}\lstinline| He attributed to a short circuit the fire which. |
\end{flushleft}\end{minipage}
\end{tabular}
\\


\midrule
4155 &
\begin{tabular}[c]{@{}l@{}}
\underline{\texttt{%
\textcolor{xkcd red}{label}:unacceptable
\textcolor{xkcd blue}{guess}:unacceptable}}
\vspace{.2em}\\
\begin{minipage}{\linewidth}\begin{flushleft}
\texttt{\textcolor{xkcd grey}{sentence}:}\lstinline| Two drops sanitize anything in your house. |
\end{flushleft}\end{minipage}
\end{tabular}
\\


\midrule
1367 &
\begin{tabular}[c]{@{}l@{}}
\underline{\texttt{%
\textcolor{xkcd red}{label}:acceptable
\textcolor{xkcd blue}{guess}:acceptable}}
\vspace{.2em}\\
\begin{minipage}{\linewidth}\begin{flushleft}
\texttt{\textcolor{xkcd grey}{sentence}:}\lstinline| We elected president the boy's guardian's employer. |
\end{flushleft}\end{minipage}
\end{tabular}
\\


\midrule
7756 &
\begin{tabular}[c]{@{}l@{}}
\underline{\texttt{%
\textcolor{xkcd red}{label}:acceptable
\textcolor{xkcd blue}{guess}:unacceptable}}
\vspace{.2em}\\
\begin{minipage}{\linewidth}\begin{flushleft}
\texttt{\textcolor{xkcd grey}{sentence}:}\lstinline| That monkey is ate the banana |
\end{flushleft}\end{minipage}
\end{tabular}
\\


\midrule
4445 &
\begin{tabular}[c]{@{}l@{}}
\underline{\texttt{%
\textcolor{xkcd red}{label}:unacceptable
\textcolor{xkcd blue}{guess}:acceptable}}
\vspace{.2em}\\
\begin{minipage}{\linewidth}\begin{flushleft}
\texttt{\textcolor{xkcd grey}{sentence}:}\lstinline| George has went to America. |
\end{flushleft}\end{minipage}
\end{tabular}
\\


\midrule
4015 &
\begin{tabular}[c]{@{}l@{}}
\underline{\texttt{%
\textcolor{xkcd red}{label}:acceptable
\textcolor{xkcd blue}{guess}:unacceptable}}
\vspace{.2em}\\
\begin{minipage}{\linewidth}\begin{flushleft}
\texttt{\textcolor{xkcd grey}{sentence}:}\lstinline| He seems intelligent to study medicine. |
\end{flushleft}\end{minipage}
\end{tabular}
\\


\bottomrule
\end{tabular}
\end{table}

\begin{table}
\centering
\caption{The Stanford Sentiment Treebank (SST2)}
\label{tab:sst2}
\begin{tabular}{cp{0.9\linewidth}}
\toprule
Index & Text
\\


\midrule
50155 &
\begin{tabular}[c]{@{}l@{}}
\underline{\texttt{%
\textcolor{xkcd red}{label}:positive
\textcolor{xkcd blue}{guess}:positive}}
\vspace{.2em}\\
\begin{minipage}{\linewidth}\begin{flushleft}
\texttt{\textcolor{xkcd grey}{sentence}:}\lstinline| a thirteen-year-old 's book report  |
\end{flushleft}\end{minipage}
\end{tabular}
\\


\midrule
58416 &
\begin{tabular}[c]{@{}l@{}}
\underline{\texttt{%
\textcolor{xkcd red}{label}:negative
\textcolor{xkcd blue}{guess}:negative}}
\vspace{.2em}\\
\begin{minipage}{\linewidth}\begin{flushleft}
\texttt{\textcolor{xkcd grey}{sentence}:}\lstinline| blues  |
\end{flushleft}\end{minipage}
\end{tabular}
\\


\midrule
59724 &
\begin{tabular}[c]{@{}l@{}}
\underline{\texttt{%
\textcolor{xkcd red}{label}:negative
\textcolor{xkcd blue}{guess}:positive}}
\vspace{.2em}\\
\begin{minipage}{\linewidth}\begin{flushleft}
\texttt{\textcolor{xkcd grey}{sentence}:}\lstinline| ` synthetic ' is the best description of this well-meaning , beautifully produced film that sacrifices its promise for a high-powered star pedigree .  |
\end{flushleft}\end{minipage}
\end{tabular}
\\


\midrule
24696 &
\begin{tabular}[c]{@{}l@{}}
\underline{\texttt{%
\textcolor{xkcd red}{label}:positive
\textcolor{xkcd blue}{guess}:negative}}
\vspace{.2em}\\
\begin{minipage}{\linewidth}\begin{flushleft}
\texttt{\textcolor{xkcd grey}{sentence}:}\lstinline| lamer instincts  |
\end{flushleft}\end{minipage}
\end{tabular}
\\


\midrule
34494 &
\begin{tabular}[c]{@{}l@{}}
\underline{\texttt{%
\textcolor{xkcd red}{label}:positive
\textcolor{xkcd blue}{guess}:positive}}
\vspace{.2em}\\
\begin{minipage}{\linewidth}\begin{flushleft}
\texttt{\textcolor{xkcd grey}{sentence}:}\lstinline| had released the outtakes theatrically and used the film as a bonus feature on the dvd  |
\end{flushleft}\end{minipage}
\end{tabular}
\\


\midrule
54555 &
\begin{tabular}[c]{@{}l@{}}
\underline{\texttt{%
\textcolor{xkcd red}{label}:negative
\textcolor{xkcd blue}{guess}:negative}}
\vspace{.2em}\\
\begin{minipage}{\linewidth}\begin{flushleft}
\texttt{\textcolor{xkcd grey}{sentence}:}\lstinline| pretentious , fascinating , ludicrous , provocative and vainglorious  |
\end{flushleft}\end{minipage}
\end{tabular}
\\


\midrule
29155 &
\begin{tabular}[c]{@{}l@{}}
\underline{\texttt{%
\textcolor{xkcd red}{label}:positive
\textcolor{xkcd blue}{guess}:negative}}
\vspace{.2em}\\
\begin{minipage}{\linewidth}\begin{flushleft}
\texttt{\textcolor{xkcd grey}{sentence}:}\lstinline| he can be forgiven for frequently pandering to fans of the gross-out comedy  |
\end{flushleft}\end{minipage}
\end{tabular}
\\


\midrule
44610 &
\begin{tabular}[c]{@{}l@{}}
\underline{\texttt{%
\textcolor{xkcd red}{label}:negative
\textcolor{xkcd blue}{guess}:positive}}
\vspace{.2em}\\
\begin{minipage}{\linewidth}\begin{flushleft}
\texttt{\textcolor{xkcd grey}{sentence}:}\lstinline| below  |
\end{flushleft}\end{minipage}
\end{tabular}
\\


\midrule
66148 &
\begin{tabular}[c]{@{}l@{}}
\underline{\texttt{%
\textcolor{xkcd red}{label}:negative
\textcolor{xkcd blue}{guess}:positive}}
\vspace{.2em}\\
\begin{minipage}{\linewidth}\begin{flushleft}
\texttt{\textcolor{xkcd grey}{sentence}:}\lstinline| 's cliche to call the film ` refreshing  |
\end{flushleft}\end{minipage}
\end{tabular}
\\


\midrule
11869 &
\begin{tabular}[c]{@{}l@{}}
\underline{\texttt{%
\textcolor{xkcd red}{label}:positive
\textcolor{xkcd blue}{guess}:negative}}
\vspace{.2em}\\
\begin{minipage}{\linewidth}\begin{flushleft}
\texttt{\textcolor{xkcd grey}{sentence}:}\lstinline| go unnoticed and underappreciated  |
\end{flushleft}\end{minipage}
\end{tabular}
\\


\midrule
55848 &
\begin{tabular}[c]{@{}l@{}}
\underline{\texttt{%
\textcolor{xkcd red}{label}:negative
\textcolor{xkcd blue}{guess}:positive}}
\vspace{.2em}\\
\begin{minipage}{\linewidth}\begin{flushleft}
\texttt{\textcolor{xkcd grey}{sentence}:}\lstinline| the film is an earnest try at beachcombing verismo , but it would be even more indistinct than it is were it not for the striking , quietly vulnerable personality of ms. ambrose .  |
\end{flushleft}\end{minipage}
\end{tabular}
\\


\midrule
57359 &
\begin{tabular}[c]{@{}l@{}}
\underline{\texttt{%
\textcolor{xkcd red}{label}:negative
\textcolor{xkcd blue}{guess}:negative}}
\vspace{.2em}\\
\begin{minipage}{\linewidth}\begin{flushleft}
\texttt{\textcolor{xkcd grey}{sentence}:}\lstinline| ( ferrera )  |
\end{flushleft}\end{minipage}
\end{tabular}
\\


\midrule
42232 &
\begin{tabular}[c]{@{}l@{}}
\underline{\texttt{%
\textcolor{xkcd red}{label}:positive
\textcolor{xkcd blue}{guess}:negative}}
\vspace{.2em}\\
\begin{minipage}{\linewidth}\begin{flushleft}
\texttt{\textcolor{xkcd grey}{sentence}:}\lstinline| forgive any shoddy product as long as there 's a little girl-on-girl action  |
\end{flushleft}\end{minipage}
\end{tabular}
\\


\midrule
15783 &
\begin{tabular}[c]{@{}l@{}}
\underline{\texttt{%
\textcolor{xkcd red}{label}:positive
\textcolor{xkcd blue}{guess}:negative}}
\vspace{.2em}\\
\begin{minipage}{\linewidth}\begin{flushleft}
\texttt{\textcolor{xkcd grey}{sentence}:}\lstinline| have finally aged past his prime ...  |
\end{flushleft}\end{minipage}
\end{tabular}
\\


\midrule
57186 &
\begin{tabular}[c]{@{}l@{}}
\underline{\texttt{%
\textcolor{xkcd red}{label}:negative
\textcolor{xkcd blue}{guess}:positive}}
\vspace{.2em}\\
\begin{minipage}{\linewidth}\begin{flushleft}
\texttt{\textcolor{xkcd grey}{sentence}:}\lstinline| hollywood war-movie stuff  |
\end{flushleft}\end{minipage}
\end{tabular}
\\


\midrule
52071 &
\begin{tabular}[c]{@{}l@{}}
\underline{\texttt{%
\textcolor{xkcd red}{label}:positive
\textcolor{xkcd blue}{guess}:negative}}
\vspace{.2em}\\
\begin{minipage}{\linewidth}\begin{flushleft}
\texttt{\textcolor{xkcd grey}{sentence}:}\lstinline| the gags  |
\end{flushleft}\end{minipage}
\end{tabular}
\\


\midrule
1896 &
\begin{tabular}[c]{@{}l@{}}
\underline{\texttt{%
\textcolor{xkcd red}{label}:positive
\textcolor{xkcd blue}{guess}:negative}}
\vspace{.2em}\\
\begin{minipage}{\linewidth}\begin{flushleft}
\texttt{\textcolor{xkcd grey}{sentence}:}\lstinline| missing from the girls ' big-screen blowout  |
\end{flushleft}\end{minipage}
\end{tabular}
\\


\midrule
64779 &
\begin{tabular}[c]{@{}l@{}}
\underline{\texttt{%
\textcolor{xkcd red}{label}:positive
\textcolor{xkcd blue}{guess}:negative}}
\vspace{.2em}\\
\begin{minipage}{\linewidth}\begin{flushleft}
\texttt{\textcolor{xkcd grey}{sentence}:}\lstinline| growing strain  |
\end{flushleft}\end{minipage}
\end{tabular}
\\


\midrule
3940 &
\begin{tabular}[c]{@{}l@{}}
\underline{\texttt{%
\textcolor{xkcd red}{label}:positive
\textcolor{xkcd blue}{guess}:negative}}
\vspace{.2em}\\
\begin{minipage}{\linewidth}\begin{flushleft}
\texttt{\textcolor{xkcd grey}{sentence}:}\lstinline| you to bite your tongue to keep from laughing at the ridiculous dialog or the oh-so convenient plot twists  |
\end{flushleft}\end{minipage}
\end{tabular}
\\




\bottomrule
\end{tabular}
\end{table}

\begin{table}
\centering
\caption{Microsoft Research Paraphrase Corpus (MRPC)}
\label{tab:mrpc}
\begin{tabular}{cp{0.9\linewidth}}
\toprule
Index & Text
\\


\midrule
799 &
\begin{tabular}[c]{@{}l@{}}
\underline{\texttt{%
\textcolor{xkcd red}{label}:equivalent
\textcolor{xkcd blue}{guess}:not equivalent}}
\vspace{.2em}\\
\begin{minipage}{\linewidth}\begin{flushleft}
\texttt{\textcolor{xkcd grey}{sentence1}:}\lstinline| We need a certifiable pay as you go budget by mid-July or schools wont open in September , Strayhorn said . |
\\
\texttt{\textcolor{xkcd grey}{sentence2}:}\lstinline| Texas lawmakers must close a $ 185.9 million budget gap by the middle of July or the schools wont open in September , Comptroller Carole Keeton Strayhorn said Thursday . |
\end{flushleft}\end{minipage}
\end{tabular}
\\


\midrule
469 &
\begin{tabular}[c]{@{}l@{}}
\underline{\texttt{%
\textcolor{xkcd red}{label}:not equivalent
\textcolor{xkcd blue}{guess}:equivalent}}
\vspace{.2em}\\
\begin{minipage}{\linewidth}\begin{flushleft}
\texttt{\textcolor{xkcd grey}{sentence1}:}\lstinline| It 's also a strategic win for Overture , given that Knight Ridder had the option of signing on Google 's services . |
\\
\texttt{\textcolor{xkcd grey}{sentence2}:}\lstinline| It 's also a strategic win for Overture , given that Knight Ridder had been using Google 's advertising services . |
\end{flushleft}\end{minipage}
\end{tabular}
\\


\midrule
1037 &
\begin{tabular}[c]{@{}l@{}}
\underline{\texttt{%
\textcolor{xkcd red}{label}:equivalent
\textcolor{xkcd blue}{guess}:not equivalent}}
\vspace{.2em}\\
\begin{minipage}{\linewidth}\begin{flushleft}
\texttt{\textcolor{xkcd grey}{sentence1}:}\lstinline| The broader Standard & Poor 's 500 Index < .SPX > edged down 9 points , or 0.98 percent , to 921 . |
\\
\texttt{\textcolor{xkcd grey}{sentence2}:}\lstinline| The Standard & Poor 's 500 Index shed 5.20 , or 0.6 percent , to 924.42 as of 9 : 33 a.m. in New York . |
\end{flushleft}\end{minipage}
\end{tabular}
\\


\midrule
1178 &
\begin{tabular}[c]{@{}l@{}}
\underline{\texttt{%
\textcolor{xkcd red}{label}:equivalent
\textcolor{xkcd blue}{guess}:not equivalent}}
\vspace{.2em}\\
\begin{minipage}{\linewidth}\begin{flushleft}
\texttt{\textcolor{xkcd grey}{sentence1}:}\lstinline| Sens. John Kerry and Bob Graham declined invitations to speak . |
\\
\texttt{\textcolor{xkcd grey}{sentence2}:}\lstinline| The no-shows were Sens. John Kerry of Massachusetts and Bob Graham of Florida . |
\end{flushleft}\end{minipage}
\end{tabular}
\\


\midrule
1753 &
\begin{tabular}[c]{@{}l@{}}
\underline{\texttt{%
\textcolor{xkcd red}{label}:equivalent
\textcolor{xkcd blue}{guess}:not equivalent}}
\vspace{.2em}\\
\begin{minipage}{\linewidth}\begin{flushleft}
\texttt{\textcolor{xkcd grey}{sentence1}:}\lstinline| The Dow Jones industrial average closed down 18.06 , or 0.2 per cent , at 9266.51 . |
\\
\texttt{\textcolor{xkcd grey}{sentence2}:}\lstinline| The blue-chip Dow Jones industrial average < .DJI > slipped 44.32 points , or 0.48 percent , to 9,240.25 . |
\end{flushleft}\end{minipage}
\end{tabular}
\\


\bottomrule
\end{tabular}
\end{table}

\begin{table}
\centering
\caption{Quora Question Pairs (QQP)}
\label{tab:qqp}
\begin{tabular}{cp{0.9\linewidth}}
\toprule
Index & Text
\\


\midrule
216515 &
\begin{tabular}[c]{@{}l@{}}
\underline{\texttt{%
\textcolor{xkcd red}{label}:duplicate
\textcolor{xkcd blue}{guess}:duplicate}}
\vspace{.2em}\\
\begin{minipage}{\linewidth}\begin{flushleft}
\texttt{\textcolor{xkcd grey}{question1}:}\lstinline| Why does Quora censor opinions and answers? |
\\
\texttt{\textcolor{xkcd grey}{question2}:}\lstinline| Does Quora censor questions and answers, and should they? |
\end{flushleft}\end{minipage}
\end{tabular}
\\


\midrule
343656 &
\begin{tabular}[c]{@{}l@{}}
\underline{\texttt{%
\textcolor{xkcd red}{label}:not duplicate
\textcolor{xkcd blue}{guess}:duplicate}}
\vspace{.2em}\\
\begin{minipage}{\linewidth}\begin{flushleft}
\texttt{\textcolor{xkcd grey}{question1}:}\lstinline| Could India's surgical strike in POK be an elaborate hoax or play? |
\\
\texttt{\textcolor{xkcd grey}{question2}:}\lstinline| Did India really conduct a surgical strike on Pakistan? |
\end{flushleft}\end{minipage}
\end{tabular}
\\


\midrule
266594 &
\begin{tabular}[c]{@{}l@{}}
\underline{\texttt{%
\textcolor{xkcd red}{label}:not duplicate
\textcolor{xkcd blue}{guess}:not duplicate}}
\vspace{.2em}\\
\begin{minipage}{\linewidth}\begin{flushleft}
\texttt{\textcolor{xkcd grey}{question1}:}\lstinline| Why is financial literacy generally not taught in American high schools? |
\\
\texttt{\textcolor{xkcd grey}{question2}:}\lstinline| Why isn't financial literacy taught in today's public schools? |
\end{flushleft}\end{minipage}
\end{tabular}
\\


\midrule
251996 &
\begin{tabular}[c]{@{}l@{}}
\underline{\texttt{%
\textcolor{xkcd red}{label}:not duplicate
\textcolor{xkcd blue}{guess}:not duplicate}}
\vspace{.2em}\\
\begin{minipage}{\linewidth}\begin{flushleft}
\texttt{\textcolor{xkcd grey}{question1}:}\lstinline| What is life like in communist countries? |
\\
\texttt{\textcolor{xkcd grey}{question2}:}\lstinline| What would life in a legitimate Communist country be like? |
\end{flushleft}\end{minipage}
\end{tabular}
\\


\midrule
7963 &
\begin{tabular}[c]{@{}l@{}}
\underline{\texttt{%
\textcolor{xkcd red}{label}:not duplicate
\textcolor{xkcd blue}{guess}:duplicate}}
\vspace{.2em}\\
\begin{minipage}{\linewidth}\begin{flushleft}
\texttt{\textcolor{xkcd grey}{question1}:}\lstinline| What is the best option for Indian politics and politicians? |
\\
\texttt{\textcolor{xkcd grey}{question2}:}\lstinline| What are the options of Indian politics and politicians? |
\end{flushleft}\end{minipage}
\end{tabular}
\\


\bottomrule
\end{tabular}
\end{table}

\begin{table}
\centering
\caption{MultiNLI (MNLI)}
\label{tab:mnli}
\begin{tabular}{cp{0.9\linewidth}}
\toprule
Index & Text
\\


\midrule
218290 &
\begin{tabular}[c]{@{}l@{}}
\underline{\texttt{%
\textcolor{xkcd red}{label}:entailment
\textcolor{xkcd blue}{guess}:neutral}}
\vspace{.2em}\\
\begin{minipage}{\linewidth}\begin{flushleft}
\texttt{\textcolor{xkcd grey}{premise}:}\lstinline| The ruins of the huge abbey of Jumiyges are perhaps the most the white-granite shells of two churches, the Romanesque Notre-Dame and the smaller Gothic Saint-Pierre. |
\\
\texttt{\textcolor{xkcd grey}{hypothesis}:}\lstinline| Notre-Dame is a larger church than Gothic Saint-Pierre. |
\end{flushleft}\end{minipage}
\end{tabular}
\\


\midrule
39431 &
\begin{tabular}[c]{@{}l@{}}
\underline{\texttt{%
\textcolor{xkcd red}{label}:neutral
\textcolor{xkcd blue}{guess}:entailment}}
\vspace{.2em}\\
\begin{minipage}{\linewidth}\begin{flushleft}
\texttt{\textcolor{xkcd grey}{premise}:}\lstinline| Unless you feel really safe in French metropolitan traffic, keep your cycling ' you can rent a bike at many railway stations ' for the villages and country roads. |
\\
\texttt{\textcolor{xkcd grey}{hypothesis}:}\lstinline| You should not cycle in the French metropolitan area.  |
\end{flushleft}\end{minipage}
\end{tabular}
\\


\midrule
27574 &
\begin{tabular}[c]{@{}l@{}}
\underline{\texttt{%
\textcolor{xkcd red}{label}:contradiction
\textcolor{xkcd blue}{guess}:neutral}}
\vspace{.2em}\\
\begin{minipage}{\linewidth}\begin{flushleft}
\texttt{\textcolor{xkcd grey}{premise}:}\lstinline| I don't think so. |
\\
\texttt{\textcolor{xkcd grey}{hypothesis}:}\lstinline| I have no real idea. |
\end{flushleft}\end{minipage}
\end{tabular}
\\


\midrule
258544 &
\begin{tabular}[c]{@{}l@{}}
\underline{\texttt{%
\textcolor{xkcd red}{label}:neutral
\textcolor{xkcd blue}{guess}:neutral}}
\vspace{.2em}\\
\begin{minipage}{\linewidth}\begin{flushleft}
\texttt{\textcolor{xkcd grey}{premise}:}\lstinline| A set of stone doors in the wall slid to the side to reveal a screen on which various torture scenes began to appear. |
\\
\texttt{\textcolor{xkcd grey}{hypothesis}:}\lstinline| The doors hid a television screen. |
\end{flushleft}\end{minipage}
\end{tabular}
\\


\midrule
320518 &
\begin{tabular}[c]{@{}l@{}}
\underline{\texttt{%
\textcolor{xkcd red}{label}:contradiction
\textcolor{xkcd blue}{guess}:neutral}}
\vspace{.2em}\\
\begin{minipage}{\linewidth}\begin{flushleft}
\texttt{\textcolor{xkcd grey}{premise}:}\lstinline| None seems comfortable with the notion of removing Clinton for sex-related misdeeds. |
\\
\texttt{\textcolor{xkcd grey}{hypothesis}:}\lstinline| People don't want Clinton touching sex related ordeals |
\end{flushleft}\end{minipage}
\end{tabular}
\\


\bottomrule
\end{tabular}
\end{table}

\begin{table}
\centering
\caption{Question NLI (QNLI)}
\label{tab:qnli}
\begin{tabular}{cp{0.9\linewidth}}
\toprule
Index & Text
\\


\midrule
1659 &
\begin{tabular}[c]{@{}l@{}}
\underline{\texttt{%
\textcolor{xkcd red}{label}:not entailment
\textcolor{xkcd blue}{guess}:not entailment}}
\vspace{.2em}\\
\begin{minipage}{\linewidth}\begin{flushleft}
\texttt{\textcolor{xkcd grey}{question}:}\lstinline| What caused Latin America's right-wing authorities to support coup o'etats? |
\\
\texttt{\textcolor{xkcd grey}{sentence}:}\lstinline| This was further fueled by Cuban and United States intervention which led to a political polarization. |
\end{flushleft}\end{minipage}
\end{tabular}
\\


\midrule
5876 &
\begin{tabular}[c]{@{}l@{}}
\underline{\texttt{%
\textcolor{xkcd red}{label}:not entailment
\textcolor{xkcd blue}{guess}:not entailment}}
\vspace{.2em}\\
\begin{minipage}{\linewidth}\begin{flushleft}
\texttt{\textcolor{xkcd grey}{question}:}\lstinline| What antenna type is a portion of the half wave dipole? |
\\
\texttt{\textcolor{xkcd grey}{sentence}:}\lstinline| The monopole antenna is essentially one half of the half-wave dipole, a single 1/4-wavelength element with the other side connected to ground or an equivalent ground plane (or counterpoise). |
\end{flushleft}\end{minipage}
\end{tabular}
\\


\midrule
5829 &
\begin{tabular}[c]{@{}l@{}}
\underline{\texttt{%
\textcolor{xkcd red}{label}:entailment
\textcolor{xkcd blue}{guess}:entailment}}
\vspace{.2em}\\
\begin{minipage}{\linewidth}\begin{flushleft}
\texttt{\textcolor{xkcd grey}{question}:}\lstinline| How are Toxicara canis infections spread? |
\\
\texttt{\textcolor{xkcd grey}{sentence}:}\lstinline| Toxocara canis (dog roundworm) eggs in dog feces can cause toxocariasis. |
\end{flushleft}\end{minipage}
\end{tabular}
\\


\midrule
77419 &
\begin{tabular}[c]{@{}l@{}}
\underline{\texttt{%
\textcolor{xkcd red}{label}:entailment
\textcolor{xkcd blue}{guess}:entailment}}
\vspace{.2em}\\
\begin{minipage}{\linewidth}\begin{flushleft}
\texttt{\textcolor{xkcd grey}{question}:}\lstinline| Why did Madrid cede the territory to the US |
\\
\texttt{\textcolor{xkcd grey}{sentence}:}\lstinline| Florida had become a burden to Spain, which could not afford to send settlers or garrisons. |
\end{flushleft}\end{minipage}
\end{tabular}
\\


\midrule
9576 &
\begin{tabular}[c]{@{}l@{}}
\underline{\texttt{%
\textcolor{xkcd red}{label}:not entailment
\textcolor{xkcd blue}{guess}:not entailment}}
\vspace{.2em}\\
\begin{minipage}{\linewidth}\begin{flushleft}
\texttt{\textcolor{xkcd grey}{question}:}\lstinline| What has no distinction between the categories  of voiced, voiceless, aspirated and unaspirated? |
\\
\texttt{\textcolor{xkcd grey}{sentence}:}\lstinline| Some of the Dravidian languages, such as Telugu, Tamil, Malayalam, and Kannada, have a distinction between voiced and voiceless, aspirated and unaspirated only in loanwords from Indo-Aryan languages. |
\end{flushleft}\end{minipage}
\end{tabular}
\\


\bottomrule
\end{tabular}
\end{table}

\begin{table}
\centering
\caption{Recognizing Textual Entailment (RTE)}
\label{tab:rte}
\begin{tabular}{cp{0.9\linewidth}}
\toprule
Index & Text
\\


\midrule
2429 &
\begin{tabular}[c]{@{}l@{}}
\underline{\texttt{%
\textcolor{xkcd red}{label}:not entailment
\textcolor{xkcd blue}{guess}:entailment}}
\vspace{.2em}\\
\begin{minipage}{\linewidth}\begin{flushleft}
\texttt{\textcolor{xkcd grey}{sentence1}:}\lstinline| Bogota, 4 May 88 - The dissemination of a document questioning Colombia's oil policy, is reportedly the aim of the publicity stunt carried out by the pro-Castro Army Of National Liberation, which kidnapped several honorary consuls, newsmen, and political leaders. |
\\
\texttt{\textcolor{xkcd grey}{sentence2}:}\lstinline| Several honorary consuls were kidnapped on 4 May 88. |
\end{flushleft}\end{minipage}
\end{tabular}
\\


\midrule
2463 &
\begin{tabular}[c]{@{}l@{}}
\underline{\texttt{%
\textcolor{xkcd red}{label}:not entailment
\textcolor{xkcd blue}{guess}:entailment}}
\vspace{.2em}\\
\begin{minipage}{\linewidth}\begin{flushleft}
\texttt{\textcolor{xkcd grey}{sentence1}:}\lstinline| The official religion is Theravada Buddhism, which is also practiced in neighboring Laos, Thailand, Burma and Sri Lanka. |
\\
\texttt{\textcolor{xkcd grey}{sentence2}:}\lstinline| The official religion of Thailand is Theravada Buddhism. |
\end{flushleft}\end{minipage}
\end{tabular}
\\


\midrule
1361 &
\begin{tabular}[c]{@{}l@{}}
\underline{\texttt{%
\textcolor{xkcd red}{label}:entailment
\textcolor{xkcd blue}{guess}:not entailment}}
\vspace{.2em}\\
\begin{minipage}{\linewidth}\begin{flushleft}
\texttt{\textcolor{xkcd grey}{sentence1}:}\lstinline| The Catering JLC formulates pay and conditions proposals of workers in the industry which, if approved by the Labour Court, legally binds employers to pay certain wage rates and provide conditions of employment. However,the QSFA now contends that the JLC has no right to make such a legally binding provision, as Section 15 of the Constitution states that the sole and exclusive power to make laws is vested in the Oireachtas, and no other authority has power to make laws for the State. It also argued that the existence of the minimum wage and 25 other pieces of legislation protecting employees' rights means that there is no need for JLCs. The chairman of the QSFA, John Grace, warned that the situation would lead to job losses and closures. |
\\
\texttt{\textcolor{xkcd grey}{sentence2}:}\lstinline| John Grace works for QSFA. |
\end{flushleft}\end{minipage}
\end{tabular}
\\






\bottomrule
\end{tabular}
\end{table}

\begin{table}
\centering
\caption{Winograd NLI (WNLI)}
\label{tab:wnli}
\begin{tabular}{cp{0.9\linewidth}}
\toprule
Index & Text
\\


\midrule
266 &
\begin{tabular}[c]{@{}l@{}}
\underline{\texttt{%
\textcolor{xkcd red}{label}:entailment
\textcolor{xkcd blue}{guess}:not entailment}}
\vspace{.2em}\\
\begin{minipage}{\linewidth}\begin{flushleft}
\texttt{\textcolor{xkcd grey}{sentence1}:}\lstinline| Susan knew that Ann's son had been in a car accident, so she told her about it. |
\\
\texttt{\textcolor{xkcd grey}{sentence2}:}\lstinline| Susan told her about it. |
\end{flushleft}\end{minipage}
\end{tabular}
\\


\midrule
478 &
\begin{tabular}[c]{@{}l@{}}
\underline{\texttt{%
\textcolor{xkcd red}{label}:entailment
\textcolor{xkcd blue}{guess}:not entailment}}
\vspace{.2em}\\
\begin{minipage}{\linewidth}\begin{flushleft}
\texttt{\textcolor{xkcd grey}{sentence1}:}\lstinline| Joe paid the detective after he delivered the final report on the case. |
\\
\texttt{\textcolor{xkcd grey}{sentence2}:}\lstinline| The detective delivered the final report on the case. |
\end{flushleft}\end{minipage}
\end{tabular}
\\


\midrule
294 &
\begin{tabular}[c]{@{}l@{}}
\underline{\texttt{%
\textcolor{xkcd red}{label}:entailment
\textcolor{xkcd blue}{guess}:not entailment}}
\vspace{.2em}\\
\begin{minipage}{\linewidth}\begin{flushleft}
\texttt{\textcolor{xkcd grey}{sentence1}:}\lstinline| Dan had to stop Bill from toying with the injured bird. He is very compassionate. |
\\
\texttt{\textcolor{xkcd grey}{sentence2}:}\lstinline| Dan is very compassionate. |
\end{flushleft}\end{minipage}
\end{tabular}
\\


\midrule
586 &
\begin{tabular}[c]{@{}l@{}}
\underline{\texttt{%
\textcolor{xkcd red}{label}:entailment
\textcolor{xkcd blue}{guess}:not entailment}}
\vspace{.2em}\\
\begin{minipage}{\linewidth}\begin{flushleft}
\texttt{\textcolor{xkcd grey}{sentence1}:}\lstinline| Dan took the rear seat while Bill claimed the front because his "Dibs!" was slow. |
\\
\texttt{\textcolor{xkcd grey}{sentence2}:}\lstinline| Dan took the rear seat while Bill claimed the front because Dan's "Dibs!" was slow. |
\end{flushleft}\end{minipage}
\end{tabular}
\\


\midrule
243 &
\begin{tabular}[c]{@{}l@{}}
\underline{\texttt{%
\textcolor{xkcd red}{label}:entailment
\textcolor{xkcd blue}{guess}:not entailment}}
\vspace{.2em}\\
\begin{minipage}{\linewidth}\begin{flushleft}
\texttt{\textcolor{xkcd grey}{sentence1}:}\lstinline| Mark was close to Mr. Singer's heels. He heard him calling for the captain, promising him, in the jargon everyone talked that night, that not one thing should be damaged on the ship except only the ammunition, but the captain and all his crew had best stay in the cabin until the work was over. |
\\
\texttt{\textcolor{xkcd grey}{sentence2}:}\lstinline| He heard Mr. Singer calling for the captain. |
\end{flushleft}\end{minipage}
\end{tabular}
\\


\bottomrule
\end{tabular}
\end{table}

\end{document}